\documentclass{article}

\PassOptionsToPackage{numbers, compress}{natbib}
\usepackage[final]{neurips_2024}
\usepackage[utf8]{inputenc} 
\usepackage[T1]{fontenc}    
\usepackage{hyperref}       
\usepackage{url}            
\usepackage{booktabs}       
\usepackage{amsfonts}       
\usepackage{nicefrac}       
\usepackage{microtype}      
\usepackage{xcolor}         

\usepackage{subfiles}
\usepackage{graphicx}
\usepackage{subcaption}
\usepackage{enumitem}
\usepackage{amsmath}
\usepackage{nameref}
\usepackage{multirow}
\usepackage{phonrule}
\usepackage{makecell}
\usepackage{wrapfig}
\usepackage{tabularx}

\newcommand{\ity}{{\textit{\mbox{-ity}}}} 
\newcommand{\ness}{{\textit{\mbox{-ness}}}}
\newcommand{\lrity}{{\textsc{\mbox{v-ity}}}}
\newcommand{\lrness}{{\textsc{\mbox{v-ness}}}}
\newcommand{\hrity}{{\textsc{\mbox{r-ity}}}}
\newcommand{\hrness}{{\textsc{\mbox{r-ness}}}}
\newcommand{\gpt}{{\mbox{GPT-J}}}

\newcommand*\samethanks[1][\value{footnote}]{\footnotemark[1]}

\usepackage{hyperref}
\hypersetup{
     colorlinks=true,
     allcolors=blue
}

\title{Derivational Morphology Reveals Analogical Generalization in Large Language Models}

\author{%
  Valentin Hofmann\thanks{Corresponding authors. E-mail: \href{mailto:valentinh@allenai.org}{\nolinkurl{valentinh@allenai.org}}; \href{mailto:janet.pierrehumbert@oerc.ox.ac.uk}{\nolinkurl{janet.pierrehumbert@oerc.ox.ac.uk}}.} \\
  Allen Institute for AI\\
  \And
  Leonie Weissweiler \\
  LMU Munich \\
  \And
  David Mortensen \\
  Carnegie Mellon University \\
  \AND
  Hinrich Sch\"utze \\
  LMU Munich, MCML \\
  \And
  Janet Pierrehumbert\samethanks \\
  University of Oxford \\
}

\begin{document}

\maketitle

\begin{abstract}
What mechanisms underlie linguistic generalization in large language models (LLMs)? This question has attracted considerable attention, with most studies analyzing the extent to which the language skills of LLMs resemble \emph{rules}. As of yet, it is not known whether linguistic generalization in LLMs could equally well be explained as the result of \emph{analogical processes}, which can be formalized as similarity operations on stored \emph{exemplars}. A key shortcoming of prior research is its focus on linguistic phenomena with a high degree of regularity, for which rule-based and analogical approaches make the same predictions. Here, we instead examine derivational morphology, specifically English adjective nominalization, which displays notable variability. We introduce a new method for investigating linguistic generalization in LLMs: we fit cognitive models that instantiate rule-based and analogical learning to the LLM training data and compare their predictions with those of the LLM, allowing us to draw direct conclusions regarding underlying mechanisms. We compare the performance of \gpt{} on a set of nonce adjectives with that of a high-performing rule-based model and a competitive analogical model. As expected, both models explain the predictions of \gpt{} equally well for adjective classes with regular nominalization patterns. However, for adjective classes with variable nominalization patterns, the analogical model provides a much better match. Furthermore, \gpt{}'s behavior is sensitive to the individual word frequencies, even for regular complex forms --- a behavior that is consistent with an analogical account of regular forms but not a rule-based one. These findings refute the hypothesis that \gpt{}'s linguistic generalization on adjective nominalization involves rules, suggesting similarity operations on stored exemplars as the underlying mechanism. Finally, we highlight a difference between linguistic generalization in humans and LLMs: while humans generalize based on types, LLMs generalize based on tokens, which we show has detrimental effects on their predictions. Overall, our study suggests that analogical processes play a bigger role in the linguistic generalization of LLMs than previously thought.
\end{abstract}

In the recent past, large language models (LLMs) such as Chinchilla \citep{hoffmann2022FIX}, Gemini \citep{geminiteam2023}, GPT-4 \citep{openai2024FIX}, LLaMA \citep{touvron2023}, Mistral \citep{jiang2023a}, OLMo \citep{groeneveld2024}, and PaLM \citep{chowdhery2023} have reached 
an unprecedented level of linguistic capability.
While some have likened the language skills of LLMs to those of humans  \citep{dale2021, haider2023}, others have
highlighted the persistent linguistic inadequacies of LLMs \citep{bender2020,dentella2023,katzir2023,weissweiler2023a}. Crucially, regardless of exactly how human-like or non-human-like the language skills of LLMs are, it is well established that they go beyond simply
copying from the training data \citep{gulordava2018,haley2020,kim2021,maudslay2021,wei2021,mccoy2023}.

What are the mechanisms underlying this kind of linguistic generalization in LLMs? Prior studies have approached this question by investigating the extent to which the language skills of LLMs resemble abstract, symbolic \emph{rules} \citep{chomsky1965,chomsky1968}. For example, the consistency with which LLMs predict the correct agreement for unseen subject-verb pairs has been interpreted as evidence that they implicitly infer a set of symbolic rules from the training data \citep{wei2021}. Comparatively speaking, much less attention has been devoted to the question of whether the language skills of LLMs could be the result of \emph{analogical processes} operating on stored \emph{exemplars}. Within cognitive science, this alternative type of generalization is argued to be a central learning mechanism and a foundation for the ability of humans to form abstract conjectures, a hallmark of human intelligence \citep{nosofsky1986,nosofsky1988,gentner2001,tenenbaum2001}. Within linguistics specifically, analogical models of word formation have proved successful in capturing detailed patterns of variation in multiple languages \citep{skousen1989,johnson1997FIX,dawdy-hesterberg2014,todd2019,ambridge2020,racz2020, racz2024}.

\begin{wraptable}{r}{0.6\linewidth}
        \caption{Key technical terms used throughout the paper.}
    \label{tab:definitions}
\footnotesize
    \centering
    \begin{tabularx}{0.6\textwidth}{lX}
    \toprule
    \textit{adjective class} & Set of adjectives ending in the same suffix.\\
     \textit{adjective nominalization} & Derivational morphology that converts adjectives to nouns.\\
     \textit{analogy} & Inference of a new word form $x$ from word forms $a$, $b$, and $c$ such that $c$ is similar to $a$, and $x$ is to $c$ as $b$ is to $a$.\\
     \textit{derivational morphology} & Operations that change the meaning or part of speech of a word.\\
     \textit{derivative} & Word form obtained by applying derivational morphology.\\
          \textit{exemplar} & A specific instance of an item that is stored in memory.\\
               \textit{frequency} & Count of a word or set of words in the corpus.\\
     \textit{nonce word} & Pseudoword invented for the purposes of an experiment.\\
     \textit{probability} & Likelihood of a form as computed by a model. \\
     \textit{rule} & Symbolic statement of a pattern.\\
    \bottomrule
    \end{tabularx}
\end{wraptable}

While LLMs store a considerable amount of their training data in their model weights \citep{biderman2023,cao2023,carlini2023,mccoy2023}, thus implicitly providing a reservoir of stored exemplars that might support analogical reasoning, it is still unclear in what ways these stored data are in fact implicated in the LLMs' language skills. As a third possibility, it could be the case that LLMs learn rules for regular linguistic phenomena while handling irregular linguistic phenomena by means of analogy over stored exemplars, in line with \emph{dual-mechanism approaches} \citep{pinker1988, pinker1991FIX, prasada1993,racz2020}.

Here, we present the first in-depth analysis of the role of analogical linguistic generalization in LLMs. Our work is motivated by a key shortcoming of the existing literature: prior studies, most of which assume rule-based generalization in LLMs \citep[e.g.,][]{wei2021}, have focused on syntactic phenomena such as subject-verb agreement, which display a high degree of regularity. Crucially, in such cases, both rule-based and analogical, exemplar-based approaches make the exact same predictions \citep{hahn1998,pothos2005,arndt-lappe2014a}; in other words, \emph{rule-like} behavior of LLMs on regular linguistic phenomena does not represent any evidence for \emph{rule-based} generalization. This very insight was at the heart of the pioneering research that first applied neural networks in the context of language learning, which argued that ``lawful behavior and judgments may be produced by a mechanism in which there is no explicit representation of the rule'' \citep{rumelhart1985}. In fact, neural network models of language depend in important ways on similarity relations amongst input examples \citep{plaut1993FIX, rumelhart1993FIX, plaut1996, plaut2000, Gonnerman.2007}, suggesting that analogy might play a major role for the language skills of LLMs. However, this hypothesis has not been systematically tested yet.

In this study, we focus on a domain of language that is known to exhibit more variability than syntax, making it better suited for distinguishing rule-based from analogical generalization: derivational morphology \citep{aronoff1976,Bauer.2001,haspelmath2010, Bauer.2013,beard2017derivation}. Specifically, we analyze how LLMs learn English adjective nominalization with \ity{} and \ness{} \citep{anshen1988FIX,baayen1996,anshen1999,lindsay2012}, 
focusing on adjectives that themselves contain a derivational suffix (e.g., \textit{avail\underline{able}}, \textit{self\underline{ish}}, \textit{hyperact\underline{ive}}). Such cases of affix stacking are an ideal testbed for our purposes since the adjective class (i.e., the adjective-final suffix) provides a controlled way to vary the regularity of the nominalization process: while some adjective classes are nominalized in a very regular way, exhibiting a clear preference for either \ity{} (e.g., adjectives ending in \textit{-able} such as \textit{available}) or \ness{} (e.g., adjectives ending in \textit{-ish} such as \textit{selfish}), others exhibit a substantial degree of variability (e.g., adjectives ending in \textit{-ive} such as \textit{hyperactive}). Furthermore, English adjective nominalization with \ity{} and \ness{} has been shown to be fully explainable as a result of analogical generalization in humans \citep{arndt-lappe2014}, suggesting that LLMs might employ the same mechanism. In general, probabilistic models \citep[][]{bresnan2007predicting}, and particularly exemplar-based analogy models \citep[][]{walsh2010, bresnan2021}, have recently proven very successful at modeling competition between nearly synonymous linguistic structures, which is an additional motivation for our work.
While there has been some previous work on the morphological capabilities of LLMs \citep[e.g.,][]{edmiston2020,haley2020,hofmann2020,hofmann2021, weissweiler2023a}, it has not diagnosed the generalization mechanisms underlying those capabilities.

As a key contribution of our work, we introduce a novel method for probing the generalization mechanisms underlying the language skills of LLMs: we fit cognitive models that instantiate certain generalization mechanisms to the LLM training data and compare their predictions on unseen data with those of the LLM. This approach, which is inspired by a long line of research in computational psychology using computer simulations \citep{sun2008, wilson2019,brasoveanu2020}, adds to the growing body of work that seeks to explain 
the behavior of LLMs as a result of the data on which they were trained \citep{wei2021,akyurek2022b,han2022,razeghi2022,elazar2023}. Our method also informs the LLM that we analyze: we choose \gpt{} \citep{wang2021a} since it is one of the few LLMs whose training data, namely the Pile \citep{gao2020}, is publicly available. For the sake of convenience, we provide short definitions of the technical terms used throughout the paper in Table~\ref{tab:definitions}.

\section*{Results}

\begin{table*}[t]
\caption{Comparison with cognitive models. The table shows real and nonce examples for the four examined adjective classes, the counts of corresponding derivatives in the Pile as well as the results of rule-based and exemplar-based analogy models evaluated against \gpt{}. The evaluation measure is accuracy. We highlight the highest accuracy value (i.e., the best-matching cognitive model) in each row in boldface --- for the two adjective classes where there is a winner (i.e., \textit{-ive} and \textit{-ous}), this is the token-based GCM model. We highlight accuracy values that are significantly ($p < .05$) worse than the highest accuracy value in each row with a $^\dag$.}
\label{tab:comparison-cognitive-modes}
\footnotesize
\centering
\begin{tabularx}{\textwidth}{llllrrrrrr}
\toprule
& & \multicolumn{2}{c}{Examples} & \multicolumn{2}{c}{Counts} & \multicolumn{2}{c}{MGL} & \multicolumn{2}{c}{GCM} \\ 
\cmidrule(lr){3-4}\cmidrule(lr){5-6}\cmidrule(lr){7-8}\cmidrule(lr){9-10}
Regularity & Suffix  & Real & Nonce & \ity{} & \ness{} & Type & Token & Type & Token\\ \midrule
\multirow{2}{*}{High} & \textit{-able} & \textit{available} & \textit{tegornable} & 11,081 & 1,034 & .893 & .893 & .893 & .893\\
& \textit{-ish} & \textit{selfish} & \textit{friquish} & 0 & 1,502 & .997 & .997 & .997 & .997\\ \midrule
\multirow{2}{*}{Low} & \textit{-ive} & \textit{sensitive} & \textit{cormasive} & 4,508 & 2,438 & $^\dag$.658 & .662 & $^\dag$.622 & \textbf{.688}\\
& \textit{-ous}  & \textit{luminous} & \textit{momogorous} & 1,372 & 2,450 & $^\dag$.657 & $^\dag$.613 & $^\dag$.610 & \textbf{.703}\\
\bottomrule
\end{tabularx}
\end{table*}

\subsection*{Generalization to Nonce Words} \label{nonce}
We compare the linguistic generalization behavior of \gpt{} with that of two high-performing cognitive models (\nameref{methods}, \nameref{methods:models}): the Minimal Generalization Learner  \citep[MGL;][]{albright2002, albright2003FIX} and the Generalized Context Model \citep[GCM;][]{nosofsky1986, nosofsky1988,nosofsky1990}. The MGL is a rule-based model that we have selected because it undertakes to capture detailed patterns of variation that earlier, simpler models did not capture. The GCM is an exemplar-based analogy model that was previously applied to variability in inflectional morphology \citep{dawdy-hesterberg2014, racz2020}. The comparison will give us some evidence about the underlying mechanism; if, for example, \gpt{} agrees consistently with the MGL over the GCM, this would suggest that it uses a similar (i.e., rule-based) mechanism under the hood.

The MGL and GCM models can be fit to either word \emph{types} or word \emph{tokens}. The inventory of word types corresponds to the list of words in a mental lexicon; only the existence of a word in the language is taken into account, and not its frequency in the training data. In an inventory of word tokens,  each occurrence of a word in the training data is treated as a separate instance, with the result that more frequent words have more instances than less frequent words.  We consider both settings, because the contrast between behaviors governed by type frequencies and those governed by token frequencies is a major theme in cognitive research on the lexicon. We focus on English adjective nominalization and examine four adjective classes (i.e., sets of adjectives ending in the same suffix), two of which clearly prefer \ity{} or \ness{}, and two of which are less regular while still showing an overall tendency towards one of the two suffixes (Table~\ref{tab:comparison-cognitive-modes}). We train the cognitive models on all adjective-derivative pairs that meet the following three criteria: (i) the adjective belongs to one of the four adjective classes in question; (ii) the derivative ends in \ity{} or \ness{}; (iii) both the adjective and the derivative occur in the Pile. For evaluation, we use UniPseudo \citep{new2023FIX} to generate 50 nonce adjectives for each of the four adjective classes (\nameref{methods}, \nameref{methods:adjectives}). We check that both the generated nonce adjective and the two corresponding derivatives 
have a frequency of zero in the Pile, i.e., they have never been seen by either the cognitive models or \gpt{}, thus providing an ideal test set for probing linguistic generalization. We then feed all nonce adjectives into the cognitive models and determine which of the two competing derivatives they prefer. For \gpt{}, we measure the probability that it assigns to the two derivatives resulting from adding \ity{} and \ness{} to the adjectives (\nameref{methods}, \nameref{methods:gptj}). Specifically, we use \mbox{\gpt{}'s} autoregressive language modeling head to compute the log probabilities for the tokens 
into which the derivatives are split by the tokenizer and sum them. We take the derivative 
with the higher total log probability as the preferred one. Since prior research has shown that varying prompts (i.e., the texts used to elicit LLM responses) can heavily affect LLM behavior \citep{rae2022FIX}, we repeat this procedure with 12 different prompts (\nameref{si}). If not stated otherwise, the presented results are averaged over prompts.

\begin{wrapfigure}{r}{0.5\linewidth}
        \centering       
    \begin{subfigure}[b]{0.21\textwidth}   
            \includegraphics[width=\textwidth]{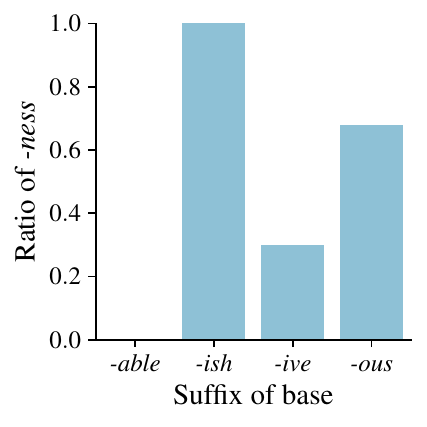}
            \vspace{-5mm}
            \caption[]%
            {{\small Type-based MGL}}    
            \label{fig:mgl-winner-ratio}
        \end{subfigure}
    \begin{subfigure}[b]{0.21\textwidth}   
            \includegraphics[width=\textwidth]{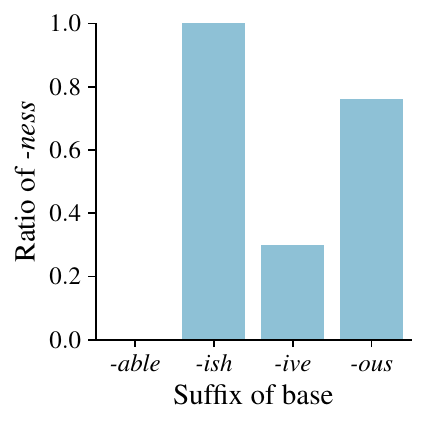}
             \vspace{-5mm}
            \caption[]%
            {{\small Token-based MGL}}    
            \label{fig:mgl-tf-winner-ratio}
        \end{subfigure}
        \par\bigskip
\begin{subfigure}[b]{0.21\textwidth}   
            \includegraphics[width=\textwidth]{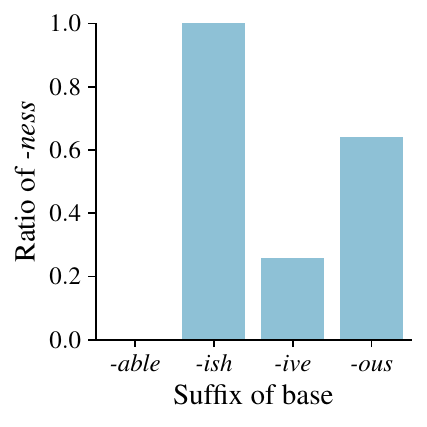}
             \vspace{-5mm}
            \caption[]%
            {{\small Type-based GCM}}    
            \label{fig:gcm-winner-ratio}
        \end{subfigure}
    \begin{subfigure}[b]{0.21\textwidth}   
            \includegraphics[width=\textwidth]{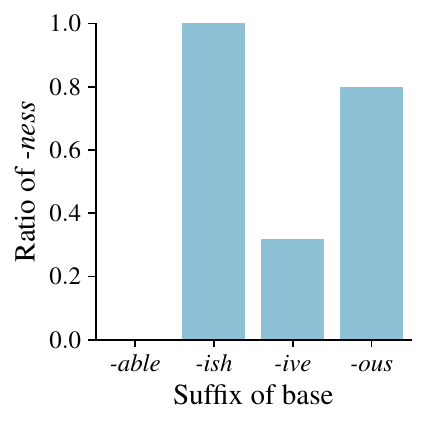}
             \vspace{-5mm}
            \caption[]%
            {{\small Token-based GCM}}    
            \label{fig:gcm-tf-winner-ratio}
        \end{subfigure}
                \par\bigskip
\begin{subfigure}[b]{0.21\textwidth}   
            \includegraphics[width=\textwidth]{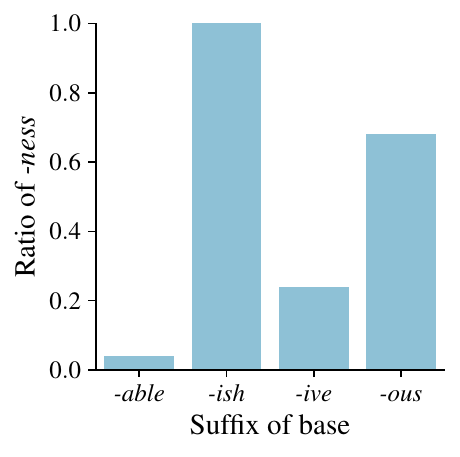}
             \vspace{-5mm}
            \caption[]%
            {{\small \gpt{}}}    
            \label{fig:gptj-winner-ratio-unseen}
        \end{subfigure}  
        \caption[]{Distribution of preferred nominalization type (specifically, ratio of \ness{} derivatives) for unseen nonce adjectives,
        for rule-based models (a, b), exemplar-based models (c, d), and \gpt{} (e). Models based on types are shown on the left (a, c), and models based on tokens are shown on the right (b, d). The ratio is computed as the number of \ness{} predictions divided by the total number of predictions (i.e., \ness{} and \ity{} predictions).
        }
        \label{fig:unseen-winners}
\end{wrapfigure}

As shown in Figure~\ref{fig:unseen-winners}, both MGL and GCM --- in the type-based as well as the token-based setting --- make completely consistent predictions for the two adjective classes that strongly prefer one affix.  They always predict \ity{} for \textit{-able} and \ness{} for \textit{-ish}. Thus, both cognitive models reproduce the regular behavior that characterizes these two adjective classes. \gpt{} also predicts \ity{} for \textit{-able}, and  it predicts \ness{} for \textit{-ish} in all but two cases for just one of the prompts (\textit{turgeishity} and \textit{prienishity}). \gpt{} is nearly as successful in capturing the regular cases as MGL and GCM, and these in turn match the predictions of \gpt{} equally well (Table~\ref{tab:comparison-cognitive-modes}, upper panel). Thus, the regular adjective classes do not tell us whether \gpt{} is more like a rule-base model or an analogical model. 

Moving to the two adjective classes that show more variability between \ity{} and \ness{} (i.e., \mbox{\textit{-ive}} and \textit{-ous}), both MGL and GCM generate variable outcomes with a higher rate of \ness{} for \textit{-ous} than for \textit{-ity}. However, the predictions differ substantially in detail:  the cognitive models (in the type-based as well as the token-based setting) agree in only 54\% of the adjective types (not shown in the figure). Crucially, the cognitive model that matches the predictions of \gpt{} on these two adjectives classes best is the token-based GCM model (Table~\ref{tab:comparison-cognitive-modes}, lower panel). As a concrete example, we consider the nonce adjective \textit{pepulative}. The MGL models map \textit{pepulative} to a rule that prescribes \ity{} following \textit{-tive}, which in the type-based as well as the token-based setting has the highest confidence of all competing rules and is hence selected by both MGL models. The GCM models, by contrast, are more strongly influenced by local similarity effects. While overall there are a larger number of \ity{} derivatives in the neighborhood of \textit{pepulative} (e.g., for adjectives ending in \textit{-lative} there are 88 derivatives with \ity{} vs.\ 27 with \ness{}), many of the adjectives particularly close to \textit{pepulative} have \ness{} derivatives with a high token frequency (e.g., \textit{manipulativeness} has a token frequency of 1,544 vs.\ 26 for \textit{manipulativity}). This difference is reflected by the GCM models, where the type-based model predicts \ity{}, but the token-based model predicts \ness{}. \gpt{}, on the other hand, prefers \ness{} for this example and hence matches the behavior of the token-based GCM model.

Our results show that the generalization behavior of LLMs on linguistic phenomena with a high degree of variability is best explained as a result of analogical mechanisms. This new finding is in line with the observation that LLMs store a considerable amount of their training data in their model weights \citep{biderman2023,cao2023,carlini2023,mccoy2023}, and it further suggests that these stored data actively contribute to the language skills displayed by LLMs. 
Our results are consistent with a model that generalizes all adjective nominalizations by analogy; they eliminate the possibility that all nominalizations are generated by rules. However, there remains the possibility that LLMs effectively use analogies in cases of variation and apply rules for adjective classes with a high degree of regularity. This possibility is suggested by earlier theories of inflectional morphology, proposing dual-mechanism models in which regular plurals and past tenses are created by rules, while irregular forms involve analogies \citep{pinker1988, pinker1991FIX, prasada1993}. To address this possibility, it is necessary to look into frequency effects for individual words, as discussed in prior work \citep{hare2001, arndt-lappe2020}. We will do so in the next sections.

\subsection*{Predictions for Seen Words}
\label{sec:stored_derivatives}

According to cognitive theories, analogies are based on remembered examples. If the mechanism underlying \gpt{}'s behavior is analogical, it must implicitly remember a large number of examples. As the first step in evaluating this inference, we ask how well \gpt{}'s behavior matches the statistics of its training data. Accurately matching the training data, derivative by derivative, would imply that the distributed representations in \gpt{} encode information about individual derivatives.

We extend the four adjective classes examined so far and include six other adjective classes that can be nominalized with either \ity{} or \ness{}: \textit{-al}, \textit{-ar}, \textit{-ed}, \textit{-ic}, \textit{-ing}, and \textit{-less}. We can divide the ten adjective classes into four groups with similar degrees of competition between \ity{} and \ness{} (see \nameref{si} for details):
\begin{itemize}
    \item \textit{-ed}, \textit{-ing}, \textit{-ish}, \textit{-less} (\hrness{}): this group 
    exhibits the highest degree of regularity and almost always takes \ness{}.
    \item \textit{-able}, \textit{-al}, \textit{-ar}, \textit{-ic} (\hrity{}): this group also exhibits a high degree of regularity (although somewhat lower than in the case of \hrness{}), with a strong tendency toward \ity{}.
    \item \textit{-ous} (\lrness{}): this adjective class exhibits a high degree of variability, with a slight tendency toward \ness{}.
    \item \textit{-ive} (\lrity{}): this adjective class also exhibits a high degree of variability, with a slight tendency toward \ity{}.
 \end{itemize}
We ask whether \gpt{} treats adjectives from these four groups differently, and whether differences between the more regular and more variable ones correspond to differences in the training data.  We draw upon the Pile and extract all derivatives ending in 
\ity{} and \ness{} whose bases belong
to one of the 10 adjective classes. To decrease noise, we only extract derivatives whose bases also occur in the Pile and apply several filtering heuristics, such as 
excluding words with non-alphabetic characters (see \nameref{si}). To include all productively formed derivatives, we do not impose a 
frequency threshold on the derivatives.

\begin{figure}[t!]
        \centering      
        \begin{subfigure}[b]{0.42\textwidth}  
            \includegraphics[width=\textwidth]{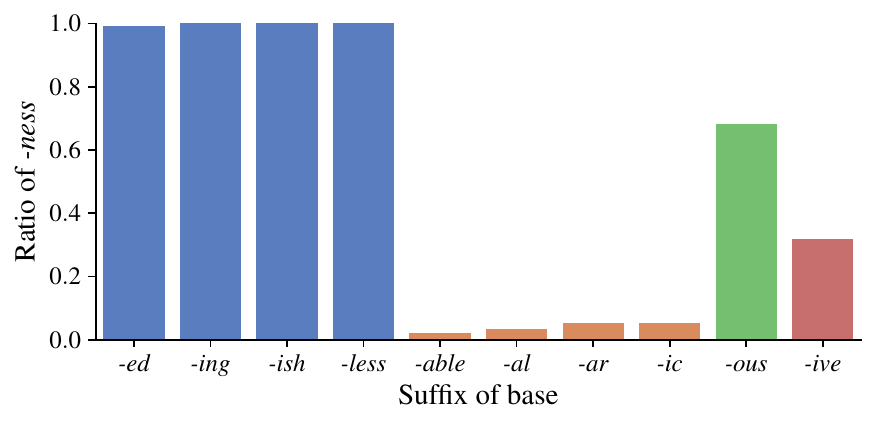}
            \vspace{-5mm}
            \caption[]%
            {{\small The Pile}}    
            \label{fig:pile-winner-ratio}
        \end{subfigure}     
        \begin{subfigure}[b]{0.42\textwidth}   
            \includegraphics[width=\textwidth]{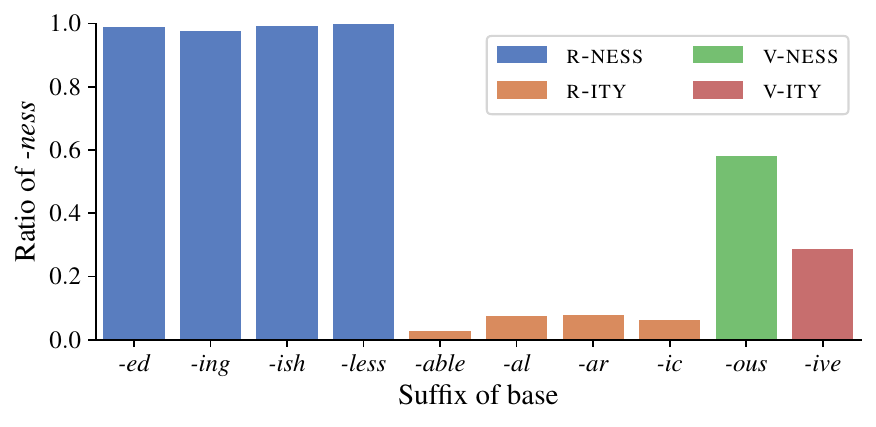}
            \vspace{-5mm}
            \caption[]%
            {{\small \gpt{}}}    
            \label{fig:gptj-winner-ratio}
        \end{subfigure}
        \caption[]{Ratio of bases preferring \ness{} in the Pile (a) and \gpt{}'s predictions with one example prompt (b). Results are similar for the other prompts. The suffixes of the base (i.e., adjective classes) are grouped by degree of competition between \ity{} and \ness{}.}
        \label{fig:pile-gptj-winner}
\end{figure}

The overall setup of probing \gpt{} is identical to the comparison with the cognitive models:
we measure the probability that \gpt{} assigns to the two derivatives
resulting from adding \ity{} and \ness{} to the adjectives, 
using the same set of prompts.
Following this procedure, we evaluate \gpt{} on all 48,995 bases from the Pile. If not stated otherwise, 
results are again averaged across prompts.

Figure~\ref{fig:pile-gptj-winner} compares, for each adjective class, the ratio of bases for which \gpt{} prefers \ness{} compared to \ity{} with the statistics from the Pile. We find that the two distributions are very similar: almost no competition for the bases in \hrness{} (i.e., \textit{-ed}, \textit{-ing}, \textit{-ish}, \textit{-less}), 
little competition for the bases in \hrity{} (i.e., \textit{-able}, \textit{-al}, \textit{-ar}, \textit{-ic}), and strong competition for \lrness{} (i.e., \textit{-ous}) and \lrity{} (i.e., \textit{-ive}). The tendency towards \ity{} and \ness{} is also exactly as predicted based on the training data --- the average correlation between the class-level \ity{}/\ness{} ratios in the training data (Figure~\ref{fig:pile-winner-ratio}) and \gpt{} predictions (Figure~\ref{fig:gptj-winner-ratio}) is 0.995 ($\pm$0.004; $p<$ 0.001 for all prompts), measured using Pearson's $r$. For multiple 
comparisons, $p$-values are corrected using the Holm-Bonferroni method \citep{Holm.1979}.

\begin{wraptable}{r}{0.4\linewidth}
\caption{Match between preferred derivatives in the training data and derivatives preferred by \gpt{}.}  
\label{tab:gptj-suffix-accuracy}
\footnotesize
\centering
\begin{tabularx}{0.4\textwidth}{llr}
\toprule
Adjective Class & Suffix & Accuracy\\ \midrule
\multirow{4}{*}{\hrness}&\textit{-ed}&.986$\pm$.007\\
&\textit{-ing}&.989$\pm$.014\\
&\textit{-ish}&.995$\pm$.004\\
&\textit{-less}&.999$\pm$.001\\
\midrule
\multirow{4}{*}{\hrity}&\textit{-able}&.896$\pm$.082\\
&\textit{-al}&.884$\pm$.073\\
&\textit{-ar}&.896$\pm$.060\\
&\textit{-ic}&.867$\pm$.090\\
\midrule
\lrness{}&\textit{-ous}&.788$\pm$.038\\
\midrule
\lrity{}&\textit{-ive}&.842$\pm$.012\\
\bottomrule
\end{tabularx}
\end{wraptable}

Furthermore, \gpt{} matches the training data statistics even on the level of individual bases: across all bases, the accuracy of \gpt{}'s preference for one of the two derivatives compared against the training data (considered here as the ground truth) is 89.5\% ($\pm$4.8\%); the derivative preferred by \gpt{} is generally the derivative that is more likely in the training data. Table~\ref{tab:gptj-suffix-accuracy} shows that there is variation between individual adjective classes, with bases 
in \hrness{} (\textit{-ed}, \textit{-ing}, \textit{-ish}, \textit{-less}) having above 95\% accuracy, 
bases in \hrity{} (\textit{-able}, \textit{-al}, \textit{-ar}, \textit{-ic}) having above 85\% accuracy, 
and bases in 
\lrity{} (\textit{-ive}) and \lrness{} (\textit{-ous}) having below 85\% accuracy, but the general level of agreement is very high.

Thus, \gpt{}'s morphological preferences closely mirror the statistics of the data it was trained on. The fact that \gpt{} very consistently prefers the derivative with the higher frequency in the training data, even in cases such as adjectives ending in \textit{-ive} where the suffix alone is a bad predictor of \ity{} vs.\ \ness{}, suggests that it stores many derivatives in its model weights. This is again in line with an analogical mechanism. However, it is still possible that some of the high-regularity adjective classes (e.g., \textit{-ish}) are handled by a rule, as suggested by dual-mechanism approaches. Next, we will disentangle these two hypotheses.

\subsection*{Frequency Effects and Analogical Pressure}
\label{sec:freq_analysis}

To test whether at least part of \gpt{}'s behavior on adjective nominalization can be explained by rules, we analyze the extent (on a log probability scale) to which \gpt{} prefers the observed nominalized form over the alternative, non-observed nominalized form. We consider only cases for which just one outcome of nominalization is attested in the Pile. The score for the unattested alternative thus represents the latent competition from the new form that might be created by rule or analogy. The difference between the two scores can be viewed as reflecting \gpt{}'s confidence in its decision to use a form that it has encountered during training. A large difference indicates high confidence, while a small difference reflects low confidence. For each adjective class, we create two sets: one in which the attested derivative has a low frequency in the Pile, $f \in (0, 10]$, and one in which the attested derivative has a high frequency in the Pile, $f \in (100, \infty)$.

If an adjective class is handled by a rule, the difference in frequency between the two sets should not affect \gpt{}'s confidence to predict the attested derivative. This is because rule-based theories abstract away from individual words; once a rule has been acquired, regular complex forms are assumed to be generated on the fly, much like complex sentence structures are, rather than being stored in memory. Rule-based theories predict that only memorized exceptions to rules will exhibit frequency effects. Many researchers might suggest the following default rule for nominalization (taking NOM to be the underlying morpheme that may be spelled out as \textit{-ness} or \textit{-ity}):
\begin{equation}\label{eq:default}
\text{NOM} \rightarrow \text{\textit{-ness}}
\end{equation}%
Under this assumption, all forms in \textit{-ity} would be memorized exceptions. 
Statistical theories of rule-formation, such as MGL, would induce the following narrower rule for \hrness{}, which has extremely strong statistical support (see 
\nameref{si} for details):
\begin{equation}\label{eq:ness}
\text{NOM} \rightarrow \text{\textit{-ness}} \: / \:
\left\{\begin{aligned}
&\text{\textit{-ed}}\\
&\text{\textit{-ing}}\\
&\text{\textit{-ish}}\\
&\text{\textit{-less}}
\end{aligned}\right\}
\: \underline{\hspace{0.4cm}}
\end{equation}%
The following subregularity for \hrity{} is also a strong candidate for status as a minor rule \citep{lakoff1970}, given the low rate of exceptions if the structural description is met.
\begin{equation}\label{eq:ity}%
\text{NOM} \rightarrow \text{\textit{-ity}} \: / \:
\left\{\begin{aligned}
&\text{\textit{-able}}\\
&\text{\textit{-al}}\\
&\text{\textit{-ar}}\\
&\text{\textit{-ic}}
\end{aligned}\right\}
\: \underline{\hspace{0.4cm}}
\end{equation}%
Empirically, the MGL model trained on the Pile contains versions of all of these rules. With respect to \gpt{}, we have already seen that the most regular outcomes \emph{can} be generated by analogy, but could they instead be generated by rule? We can address this question by probing whether the cases that fall under rules \ref{eq:default}, \ref{eq:ness}, or \ref{eq:ity} exhibit word frequency effects on \gpt{}'s confidence in its decision. If a rule is active for a group of cases, its outputs should be exempt from word frequency effects since the outputs --- and hence their individual frequencies --- are not stored. 

\begin{wrapfigure}{r}{0.5\linewidth}
\centering        
\includegraphics[width=0.48\textwidth]{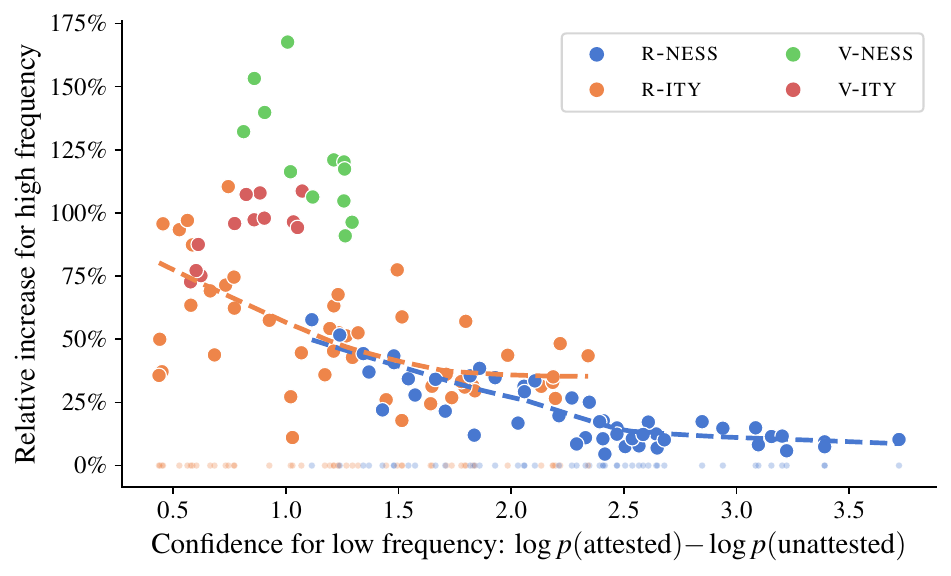}
\caption{Impact of word frequency on \gpt{}'s confidence in its choice. 
\mbox{x-axis}: Log probability difference between the attested and unattested choices for low-frequency derivatives with $f \in (0, 10]$.  We have converted the log probabilities from base $e$ to base $10$ for better readability. y-axis: Relative increase in confidence for high-frequency derivatives with $f \in (100, \infty)$. 
Each dot corresponds to \gpt{}'s predictions for an adjective class given a specific prompt. Dots are colored by degree of competition between \ity{} and \ness{}. We added LOWESS lines for \hrness{} and \hrity{}. Dots at $y= 0\%$ indicate the expected behavior if \hrness{} and \hrity{} were handled by rule.}
\label{fig:linear-regression-summary}
\end{wrapfigure}

Figure \ref{fig:linear-regression-summary} shows \gpt{}'s confidence (i.e., 
the natural log probability difference between the attested and the unattested derivative) 
for low-frequency derivatives with $f \in (0, 10]$ vs.\ the relative increase in confidence for high-frequency derivatives with $f \in (100, \infty)$ for all adjective classes, divided into the four regularity-based groups. The relative increase in confidence is positive for all adjective classes and for all prompts, indicating that \gpt{} is always more confident in its decision for the frequent than the rare derivatives, even for the \hrness{} class. This indicates that the model has stored distributed representations for \emph{all} the derivatives, contrary to the predictions of the dual-mechanism model. Put differently, none of the adjective classes are handled by rule.

Given that the cognitive models considered above (i.e., MGL and GCM) generate no examples of \textit{-ishity} (reflecting the non-occurrence of such examples in the training data), it is striking that \gpt{} exhibited any uncertainty at all about these cases. Recall, however, that it made actual errors as well (see \nameref{nonce}). This outcome can be attributed to the model paying some attention to parts of the word that precede the suffix of the base (e.g., \textit{-ish}), even though these parts have no linguistic relevance to the form of the nominalization. Note that transformer-based language models such as \gpt{} lack the ability to pay zero attention to any part of the input.

Overall, the examined adjective classes exhibit a downward slope in Figure \ref{fig:linear-regression-summary}, which is also reflected by the \hrness{} and \hrity{} groups individually (indicated by trendlines). Thus, the more confident the model was in its decisions for the low-frequency group, the smaller the effect of word frequency on confidence. This finding is difficult to explain under the assumption of rules, but we will show that it is perfectly in line with analogy as the underlying generalization mechanism. More specifically, we will attribute the downward slope to variation in the \emph{analogical pressure} for the different adjective classes. It is well established that frequent words have more robust representations in memory than rare words. This generalization  plays a central role in our understanding of how analogy functions in morphology. During language production, analogical processes supply a complex word form if the form has never been encountered during learning. But, they also exert pressure on known forms stored in memory. The better the complex form has been learned as such, the less susceptible it is to these pressures and the more quickly and reliably it is retrieved in its remembered form. This mechanism operating over multiple generations is the one by which frequent forms can retain exceptional inflectional patterns while rare forms are regularized \citep{corbett2001, bybee2006}; for example, English \textit{say, take, come} still retain their exceptional past tense forms (\textit{said, took, came}) while previously irregular rare verbs, such as \textit{writhe, flay, spew} do not \citep{lieberman2007}. A Bayesian model that formalizes analogical pressure, including the frequency effect just described, is developed in \citep{daland2007new}.

For the most regular \hrness{} adjective classes, there is little analogical pressure on any specific target form because there are few if any examples that have the suffix followed by \mbox{\textit{-ity}}. \gpt{} has high confidence in its answer, and the weak analogical pressure comes from a few forms that happen to have similarities in the stem. Word frequency, it appears, has a relatively small effect on the confidence when the confidence is already high. Lower confidence levels for the low-frequency forms towards the left of the graph represent cases in which the neighborhood of the target form is more heterogeneous. With increased competition, we observe that the effect of word frequency becomes greater. We quantify this by calculating the Shannon entropy of the distribution over \ity{} and \ness{} as the preferred form in the Pile, and using Pearson's $r$ as a measure of the correlation with the confidence increase for high-frequency derivatives. At $r^2 = 0.75$, $p < 0.001$, this correlation is highly significant. Thus, the heterogeneity of the exemplar neighborhood for an adjective class predicts the extent to which \gpt{}'s confidence relies on frequency --- a finding that is exactly in line with the predictions of models assuming analogical pressure \citep[e.g.,][]{daland2007new} while being completely at odds with rule-based approaches, which do not assume such frequency effects to begin with.

The left side of Figure \ref{fig:linear-regression-summary} exhibits more variability than the right side. We believe that this variability is caused by local neighborhood effects and the interaction of these effects with the prompting mechanism. Recall from the discussion of the nonce word \textit{pepulative} that analogical models are sensitive not merely to the overall statistics for the two competing nominalizations, but also to the similarity and frequency of the most similar neighbors (see \nameref{nonce}). These localized effects --- which for the case of attested derivatives would also include semantic similarity --- create a lumpy prediction landscape whose properties we do not try to quantify here. Meanwhile, the prompting mechanism is known to influence the focus and bias of the underlying transformer model \citep{petrov2024}. Slightly different prompts direct the focus towards different parts of the lumpy landscape, and would hence produce noise in the datapoints for Figure \ref{fig:linear-regression-summary}.

To sum up, our analysis suggests that \gpt{} learns adjective nominalization by implicitly storing derivatives in its model weights. In cases where the exemplar neighborhood is highly homogeneous, \gpt{} produces highly regular outputs, but nonetheless reveals evidence of analogical pressure in its confidence scores. While regular, or rule-like, behavior of LLMs has been observed before \citep[e.g.,][]{wei2021}, our results contextualize this finding in important ways, suggesting that rule-like behavior forms the end of a gradient characterized by varying levels of regularity. This result is not consistent with assuming a qualitative difference between forms derived by rule and stored exemplars. It is exactly in line with the predictions of exemplar-based analogy models \citep[e.g.,][]{dawdy-hesterberg2014, racz2020, racz2024}.

\subsection*{Human Use of Word Types Versus Tokens}

We have established that \gpt{} relies on token-level analogical generalization. In contrast, previous studies have concluded that humans generalize over word types \citep{bybee1995, pierrehumbert2001, pierrehumbert2003}: their propensity to generalize a word formation pattern depends on the number of distinct word types in the individual's mental lexicon that support the pattern (referred to as the size of the lexical gang). This points to a difference between the morphological processing in humans and LLMs. We will now investigate this difference in greater detail, by comparing the predictions of \gpt{} to judgments made by humans. 

\subsubsection*{\textit{Judgments of Nonce Words}}

First, we make a direct comparison to \gpt{}'s behavior for nonce words. 
22 native English speaker volunteer annotators indicated their preference for the \ity{} versus the \ness{} derivative of each nonce adjective in our study (\nameref{si}). Because \gpt{} is not a state-of-the-art model, we also introduce an additional comparison, by asking whether a more recent model is more human-like in its judgments. Specifically, we evaluate \mbox{GPT-4} \citep{openai2024FIX} on the same set of adjectives (see \nameref{methods}, \nameref{methods:gpt4} for implementation details). If \mbox{GPT-4} displays more human-like judgments than \gpt{}, then the trend of improving LLMs through larger training sets and bigger model sizes will have paid off in this domain.

\begin{wraptable}{r}{0.6\linewidth}
\caption{Human evaluation. The table shows the results of rule-based and exemplar-based analogy models as well as \gpt{} and GPT-4 evaluated against human annotations. The measure is accuracy.}  
\label{tab:comparison-humans}
\footnotesize
\centering
\begin{tabularx}{0.6\textwidth}{lrrrrrr}
\toprule
 & \multicolumn{2}{c}{MGL} & \multicolumn{2}{c}{GCM} & \multicolumn{2}{c}{LLMs}\\ 
\cmidrule(lr){2-3}\cmidrule(lr){4-5}\cmidrule(lr){6-7}
Suffix  & Type & Token & Type & Token & \gpt{} & GPT-4 \\ \midrule
\textit{-able} & 1.000 & 1.000 & 1.000 & 1.000 & .893 & .960 \\
\textit{-ish} & 1.000 & 1.000 & 1.000 & 1.000 & .997 & 1.000 \\
\textit{-ive} & .720 & .680 & .760 & .700 & .632 & .440 \\
\textit{-ous} & .560 & .520 & .640 & .520 & .503 & .400\\
\bottomrule
\end{tabularx}
\end{wraptable}

In Table \ref{tab:comparison-humans}, we take the derivative (with \ity{} or \ness{}) more often selected by humans as the ground truth. The table
gives the accuracy of \gpt{}, \mbox{GPT-4}, as well as the cognitive models considered above (i.e., MGL and GCM), measured against this human response. The type-based GCM model overall matches the human behavior best. While all cognitive models perfectly reproduce the homogeneous behavior for \textit{-able} and \textit{-ish}, the type-based GCM model better matches the human predictions for \textit{-ive} and \textit{-ous}, as reflected by large gaps compared to the second best model, type-level MGL (\textit{-ive}: 4\%, \textit{-ous}: 8\%). The token-based variants of MGL and GCM match the human behavior substantially worse than the type-based variants, which is exactly in line with what has been suggested in prior work \citep{pierrehumbert2001, albright2003FIX}. Moving to the results for \gpt{}, it turns out to  match the human responses worse than any of the cognitive models, for all four adjective classes. The gap compared to the best cognitive model, type-based GCM, is considerable, especially for \textit{-ive} and \textit{-ous}, amounting to roughly 13\% in both cases. The picture is overall even \emph{worse} for \mbox{GPT-4}. While the predictions for the high-regularity classes are good (almost always \ity{} for \textit{-able} and \ness{} for \textit{-ish}, like all other models), the match with humans is more than 10\% worse than \gpt{} for both \textit{-ive} and \textit{-ous}.

Why do \gpt{} and \mbox{GPT-4} match the human behavior so much worse than the much simpler type-based GCM? The key factor, we argue, is that both of these LLMs are driven by the token frequencies of the words in the training data.  Just as for \gpt{}, the token-based GCM and MGL models match the behavior of \mbox{GPT-4} better than the type-based models (see \nameref{si}).  Token-oriented behavior is desirable in that it results in highly realistic implicit knowledge of individual words, as we have seen above. However, humans step back from the frequencies of individual words when making generalizations about possible words. LLMs seem to lack the ability to do this.

Furthermore, our results suggest that \mbox{GPT-4's} over-reliance on token frequency is if anything worse than that of  \gpt{}. Thus \mbox{GPT-4's} morphological generalization behavior seems to be even \emph{less} human-like than that of \gpt{}. This finding  is reminiscent of recently reported `inverse scaling effects', more specifically the tendency of larger LLMs to rely even more strongly on prior statistics from the training data than smaller models do \citep{mckenzie2023}.

Since the performance of the best model, the type-based GCM, leaves room for improvement, we can ask why its performance was not better. The single biggest discrepancy was that the GCM selected \textit{-ness} after \textit{-ous} more than the humans did. Our analysis does not deal with the possibility that some people may consider the affix \textit{-osity} to be a unified affix bundle, along the lines suggested in \citep{Stump.2017, Stump.2019}; this would enhance its availability. Given that the GCM was fit to all word pairs attested in the Pile, the analysis also failed to allow for differences amongst human mental lexicons (whether in the entries themselves, or in the extent to which morphological relations between entries have been adduced). Other studies have found considerable variability amongst human participants in the area of derivational morphology in general, and in preference for \textit{-ity} over \mbox{\textit{-ness}} specifically \citep{Pierrehumbert.2006FIX, saily2011FIX, saily2014, saily2016, pierrehumbert2016FIX}. In this context, it is noteworthy that all participants of our annotation study held at least a college degree (see \nameref{si} for details), whereas much of the Pile consists of web data such as informal discussions on Reddit or StackExchange \citep{gao2020}. There is thus the possibility that there was a misalignment between the sociolect most strongly represented in the Pile and the ideolects sampled as part of our annotation study. Finally, the GCM works on the basis of word forms only, and has no way of taking into account similarities in meaning that also play a role in shaping morphological systems. In contrast, LLMs like \gpt{} and GPT-4 do have the ability to consider similarities in meaning, but any advantage they may gain from their semantics appears to be more than offset by the drawbacks of their reliance on token frequencies.  

\subsubsection*{\textit{Familiarity of Complex Words}}

Our results on nominalizations indicate that \gpt{} and \mbox{GPT-4} do not have a mental lexicon in the sense that humans do, in that they lack the ability to step back from word tokens and generalize over word types. Here, we present a brief demonstration that this observation pertains to morphologically complex words more generally, and not just to nominalizations. For this demonstration, we draw on the Hoosier Lexicon, a dataset of 19,320 English words that includes word frequencies and familiarity ratings on a seven-point Likert Scale \citep{nusbaum1984FIX}. 
An important finding of the original study was a dissociation between word frequency and rated familiarity; one might expect the two to be highly correlated, however some infrequent words are judged as much more familiar than their frequency would suggest. \citet{needle2022} identify morphological structure as an important factor contributing to this dissociation. A word like \textit{precancellation}, with a recognizable prefix, stem, and affix seems familiar even though it is rare, on the strength of the familiarity of its parts. 

We analyze the $n=2,835$ words in the Hoosier lexicon that have a frequency of less than $10,000$ in the Pile (corresponding to a frequency of roughly $1$ in 50,000,000 words or less). Leveraging the CELEX dictionary \citep{baayen1996} and methodology from prior work \cite{hofmann2020a, hofmann-etal-2020-graph}, we use affix-stripping to identify the $n=1,005$ words that exemplify derivational morphology by virtue of being parsable as a simpler word plus any combination of affixes. $n=1,830$ words cannot be parsed in this way, and we consider them to be simplex words (see \nameref{si}). For human judgments, we take the familiarity ratings reported by \citet{nusbaum1984FIX}. We estimate the `familiarity' that \gpt{} assigns to a word as  the log probability that it assigns in the context of neutral prompts (see \nameref{methods}, \nameref{methods:vocabulary} for details). 
Comparing log probabilities to human familiarity ratings is justified because the probabilities assigned to words by language models are known to correlate with psycholinguistic measures of lexical access \citep[e.g., reading times;][]{wilcox2023}, which for humans are impacted by familiarity to a larger extent than frequency \citep{brysbaert2019}.

\begin{wrapfigure}{r}{0.5\linewidth}
        \centering      
        \begin{subfigure}[b]{0.21\textwidth}  
            \includegraphics[width=\textwidth]{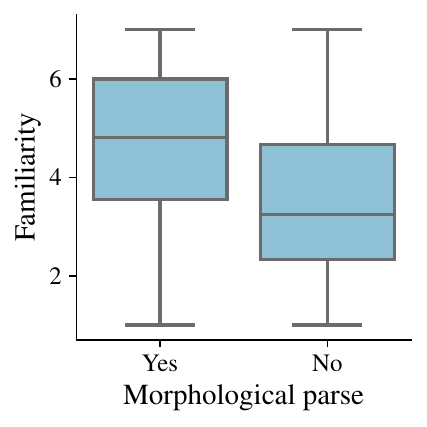}
            \vspace{-5mm}
            \caption[]%
            {{\small Humans}}    
            \label{fig:low-frequency-humans}
        \end{subfigure}     
        \begin{subfigure}[b]{0.21\textwidth}   
            \includegraphics[width=\textwidth]{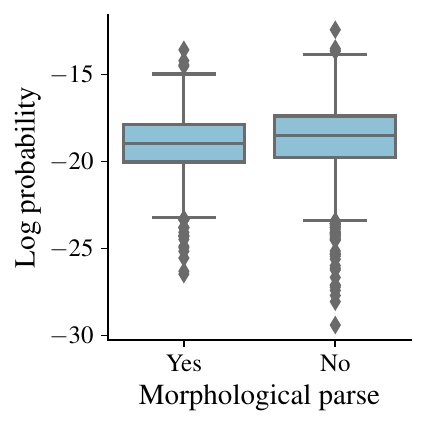}
            \vspace{-5mm}
            \caption[]%
            {{\small \gpt{}}}    
            \label{fig:low-frequency-gptj}
        \end{subfigure}
        \caption[]{Impact of morphological decomposability of words on their familiarity as rated by human annotators (a) and the log probability assigned to them by \gpt{} (b).}
        \label{fig:low-frequency}
\end{wrapfigure}

Results for humans are displayed in Figure \ref{fig:low-frequency-humans}. 
The average familiarity of words with a morphological parse ($n=1,005$) is significantly higher than the average familiarity of words with no morphological parse ($n=1,830$), $t(2120.2) = 19.2$, $p < .001$ by a Welch's $t$-test. This confirms the results reported by \citet{needle2022}. Due to this important factor, the correlation between familiarity and log frequency in the entire Hoosier lexicon proves to be modest according to a linear regression, $F(1, 19318) = 11251.2$, $R^2 = 0.368$, $p < .001$. For \gpt{}, on the other hand, words with a morphological parse do not have any advantage (Figure \ref{fig:low-frequency-gptj}); quite the opposite, the estimated familiarity of words with a morphological parse is significantly \emph{lower} than the estimated familiarity of words with no morphological parse for \gpt{}, $t(2285.9) = -4.9$, $p < .001$. This outcome can be explained by the fact that the correlation between the log frequencies and the log probabilities assigned to the words by \gpt{} is very high, $F(1, 19318) = 58553.5$, $R^2 = 0.752$, $p < .001$, and the target words without a parse have somewhat higher average frequency ($m = 4285.1$) than those having a parse ($m = 4093.7$).  

Experimental studies on wordlikeness judgments \cite{needle2022} and on speech perception \cite{Rastle.2004, Taft.2004, Beyersmann.2015} show that humans continually monitor for known words inside rare or novel words. 
This means that their type-level lexical representations are exploited during processing, and can cause rare words to seem familiar. \gpt{} does not rely on this type-level mechanism and hence lacks the dissociation between frequency and familiarity that is caused by morphological structure.  

\section*{Discussion}

This paper provides the first empirical evidence for analogical linguistic generalization in LLMs. We found that an analogical cognitive model best explains how \gpt{} nominalizes unseen nonce adjectives whose adjective class exhibits a high degree of variability. While the analogical and the rule-based model explained \gpt{}'s predictions on adjective classes with a high degree of regularity equally well, we showed that the rule-like behavior for those adjectives is the end point of a continuum, where the position of an adjective class is precisely predictive from the level of heterogeneity in the training data. This result is in line with the predictions made by exemplar-based analogy models \citep[e.g.,][]{todd2019}, and it is not consistent with assuming rules, not even with assuming rules for part of the adjective classes as in dual-mechanism approaches. We further found that \gpt{} stores a considerable amount of seen derivatives in its weights, again in line with analogical generalization.

Humans have been also argued to employ analogical generalization in adjective nominalization. However, while humans generalize based on types, we showed that \gpt{} generalizes based on tokens. Is this difference between humans and LLMs reflected by their predictions? We found that this is indeed the case: the predictions of \gpt{}, and similarly of GPT-4, are less human-like than any of the examined cognitive models. We further found that a central manifestation of type-level representations in humans, specifically decomposition into morpheme types, is not mirrored by language models. This suggests a critical difference in the organization of the lexicon between humans and LLMs: while the mental lexicon of humans is organized around types, the lexicon of LLMs is organized around tokens. Given that analogical generalization mechanisms in central ways depend on the lexicon they are operating on, a non-human-like lexicon leads to non-human-like generalizations, which is confirmed by our experiments.

Our findings are also related to the ongoing debate about how human-like the language skills of LLMs are \cite{dentella, hu2023prompting, hu2024language}, by highlighting a clear example of an area where the generalizations of LLMs --- even the most performant ones --- are decidedly non-human-like. The specific shortcoming that we observe in LLMs is that they do not distill token occurrences in text into more abstract type-level representations. From a semantics perspective, this can be interpreted as a failure to semantically ascend \citep{quine1960} to a level of representations that would make it possible for the LLMs to generalize over linguistic objects (specifically, word relations). More generally, this finding can be connected to converging evidence that LLMs fail to form meta-level representation the way humans do \citep{dziri2023}, in our case meta-level \emph{linguistic} representations.

\section*{Materials and Methods} \label{methods}

\subsection*{Cognitive Models}
\label{methods:models}

MGL \citep{albright2002, albright2003FIX} works by inferring abstract rules from the lexicon. It starts by iterating over pairs of words and forming initial generalizations based on shared phonological features, which are then iteratively merged, yielding increasingly abstract rules. Each rule is associated with a value signifying its accuracy in the lexicon, potentially adjusted by its overall support in the lexicon. To make predictions for a new input, the rule with the highest accuracy that matches the phonological properties of the input is selected. 
GCM \citep{nosofsky1986, nosofsky1988,nosofsky1990} does not infer abstract rules but instead stores all forms from the training data in an inventory. To make predictions for a new input, the input is compared to all instances that exhibit a specific type of output (e.g., to all bases that have a derivative with \ity{}). The type of output with the highest total similarity to the input is selected.

We use the implementation of MGL made available by Albright and Hayes \cite{albright2003FIX}, using default hyperparameters. For GCM, our implementation exactly follows prior studies in linguistics using the model \citep[e.g.,][]{albright2003FIX,dawdy-hesterberg2014,racz2020}.

\subsection*{Nonce Adjectives}
\label{methods:adjectives}

To create the nonce adjectives for the four adjective classes, we draw upon UniPseudo 
\citep{new2023FIX}. UniPseudo uses an algorithm based on Markov chains of orthographic $n$-grams
that it applies to a specifiable list of input words, generating a 
list of pseudowords. Importantly, 
when all input words end in a certain sequence of characters, the generated pseudowords also end in that sequence of
characters. We leverage this property of UniPseudo to generate 50 nonce adjectives for each adjective class based on a curated list of adjectives drawn from CELEX \citep{baayen1995}, MorphoLex \citep{sanchez-gutierrez2018}, and MorphoQuantics \citep{laws2014}. For pseudoword length, we use the two most frequent lengths as measured on the extracted adjectives for each class and generate 25 pseudowords for each length. We use the bigram algorithm. See \nameref{si} for the full list of pseudowords.

\subsection*{\gpt{}}
\label{methods:gptj}

We describe the method we use to probe \gpt{} more formally.
Let $b$ be a base (e.g., \textit{sensitive}) and $s$ be a suffix (e.g., \ity{}). We denote with $d(b, s)$ the derivative resulting from adding $s$ to $b$ and applying all required morpho-orthographic changes (e.g., deletion of base-final \textit{e}). For instance, for $b = \text{\textit{sensitive}}$ and $s=\text{\ity{}}$, we have $d(b, s) = \text{\textit{sensitivity}}$. To measure the probability that \gpt{} assigns to $d(b,s)$ as a derivative of $b$, 
we use various prompts $t(b)$. While some of the prompts ask \gpt{} to nominalize $b$ (e.g., $t(b) = \text{\textit{Turn the given adjective into a noun. }} b \rightarrow$), others are less explicit (e.g., 
$t(b) = b \rightarrow$). See \nameref{si} for the full set of prompts.

Given a filled prompt $t(b)$, we pass it through \gpt{} and measure the probability that \gpt{} assigns to the two derivatives $d(b, \text{\ity{}})$ and $d(b, \text{\ness{}})$ as continuations of $t(b)$. 

We use the \gpt{} implementation available on Hugging Face \citep{wolf2020}. \gpt{} has a total of 6,053,381,344 parameters.
All experiments are performed on a stack of eight GeForce
GTX 1080 Ti GPUs (11GB).

\subsection*{GPT-4}
\label{methods:gpt4}

Since the OpenAI API does not provide access to the token-level probabilities, we cannot use the same method as for \gpt{}. Instead, we leverage GPT-4 instruction-following capabilities and directly ask it which of the two derivatives it prefers for a given nonce adjective.

\subsection*{Vocabulary Test}
\label{methods:vocabulary}

 To measure the probability that \gpt{} assigns to a word $w$, we use various prompts $t(w)$ (e.g., $t(w) = \text{\textit{The following is a word: }} w$). See \nameref{si} for the full set of prompts.
 
 Given a filled prompt $t(w)$, we pass it through \gpt{} and measure the probability that \gpt{} assigns to the word.

\section*{Acknowledgments}

This work was funded by the Engineering and Physical Sciences Research Council (grant EP/T023333/1 awarded to University of Oxford). Valentin Hofmann was also supported by the German Academic Scholarship Foundation.

{\small
\bibliography{main}

\begin{thebibliography}{120}
\providecommand{\natexlab}[1]{#1}
\providecommand{\url}[1]{\texttt{#1}}
\expandafter\ifx\csname urlstyle\endcsname\relax
  \providecommand{\doi}[1]{doi: #1}\else
  \providecommand{\doi}{doi: \begingroup \urlstyle{rm}\Url}\fi

\bibitem[Hoffmann et~al.(2022)Hoffmann, Borgeaud, Mensch, Buchatskaya, Cai,
  Rutherford, de~Las~Casas, Hendricks, Welbl, Clark, Hennigan, Noland,
  Millican, van~den Driessche, Damoc, Guy, Osindero, Simonyan, Elsen, Rae,
  Vinyals, and Sifre]{hoffmann2022FIX}
Jordan Hoffmann, Sebastian Borgeaud, Arthur Mensch, Elena Buchatskaya, Trevor
  Cai, Eliza Rutherford, Diego de~Las~Casas, Lisa~Anne Hendricks, Johannes
  Welbl, Aidan Clark, Tom Hennigan, Eric Noland, Katie Millican, George van~den
  Driessche, Bogdan Damoc, Aurelia Guy, Simon Osindero, Karen Simonyan, Erich
  Elsen, Jack~W. Rae, Oriol Vinyals, and Laurent Sifre.
\newblock Training {C}ompute-{O}ptimal {L}arge {L}anguage {M}odels.
\newblock Preprint, arXiv 2203.15556, 2022.

\bibitem[Team(2023)]{geminiteam2023}
Gemini Team.
\newblock Gemini: {{A Family}} of {{Highly Capable Multimodal Models}}.
\newblock Preprint, arXiv 2312.11805, 2023.

\bibitem[OpenAI(2024)]{openai2024FIX}
OpenAI.
\newblock {GPT-4} {T}echnical {R}eport.
\newblock Preprint, arXiv 2303.08774, 2024.

\bibitem[Touvron et~al.(2023)Touvron, Lavril, Izacard, Martinet, Lachaux,
  Lacroix, Rozi{\`e}re, Goyal, Hambro, Azhar, Rodriguez, Joulin, Grave, and
  Lample]{touvron2023}
Hugo Touvron, Thibaut Lavril, Gautier Izacard, Xavier Martinet, Marie-Anne
  Lachaux, Timoth{\'e}e Lacroix, Baptiste Rozi{\`e}re, Naman Goyal, Eric
  Hambro, Faisal Azhar, Aurelien Rodriguez, Armand Joulin, Edouard Grave, and
  Guillaume Lample.
\newblock {{LLaMA}}: {{Open}} and {{Efficient Foundation Language Models}}.
\newblock Preprint, arXiv 2302.13971, 2023.

\bibitem[Jiang et~al.(2023)Jiang, Sablayrolles, Mensch, Bamford, Chaplot,
  de~las Casas, Bressand, Lengyel, Lample, Saulnier, Lavaud, Lachaux, Stock,
  Scao, Lavril, Wang, Lacroix, and Sayed]{jiang2023a}
Albert~Q. Jiang, Alexandre Sablayrolles, Arthur Mensch, Chris Bamford,
  Devendra~Singh Chaplot, Diego de~las Casas, Florian Bressand, Gianna Lengyel,
  Guillaume Lample, Lucile Saulnier, L{\'e}lio~Renard Lavaud, Marie-Anne
  Lachaux, Pierre Stock, Teven~Le Scao, Thibaut Lavril, Thomas Wang,
  Timoth{\'e}e Lacroix, and William~El Sayed.
\newblock Mistral {{7B}}.
\newblock Preprint, arXiv 2310.06825, 2023.

\bibitem[Groeneveld et~al.(2024)Groeneveld, Beltagy, Walsh, Bhagia, Kinney,
  Tafjord, Jha, Ivison, Magnusson, Wang, Arora, Atkinson, Authur, Chandu,
  Cohan, Dumas, Elazar, Gu, Hessel, Khot, Merrill, Morrison, Muennighoff, Naik,
  Nam, Peters, Pyatkin, Ravichander, Schwenk, Shah, Smith, Strubell, Subramani,
  Wortsman, Dasigi, Lambert, Richardson, Zettlemoyer, Dodge, Lo, Soldaini,
  Smith, and Hajishirzi]{groeneveld2024}
Dirk Groeneveld, Iz~Beltagy, Pete Walsh, Akshita Bhagia, Rodney Kinney, Oyvind
  Tafjord, Ananya~Harsh Jha, Hamish Ivison, Ian Magnusson, Yizhong Wang, Shane
  Arora, David Atkinson, Russell Authur, Khyathi~Raghavi Chandu, Arman Cohan,
  Jennifer Dumas, Yanai Elazar, Yuling Gu, Jack Hessel, Tushar Khot, William
  Merrill, Jacob Morrison, Niklas Muennighoff, Aakanksha Naik, Crystal Nam,
  Matthew~E. Peters, Valentina Pyatkin, Abhilasha Ravichander, Dustin Schwenk,
  Saurabh Shah, Will Smith, Emma Strubell, Nishant Subramani, Mitchell
  Wortsman, Pradeep Dasigi, Nathan Lambert, Kyle Richardson, Luke Zettlemoyer,
  Jesse Dodge, Kyle Lo, Luca Soldaini, Noah~A. Smith, and Hannaneh Hajishirzi.
\newblock {{OLMo}}: {{Accelerating}} the {{Science}} of {{Language Models}}.
\newblock Preprint, arXiv 2402.00838, 2024.

\bibitem[Chowdhery et~al.(2023)Chowdhery, Narang, Devlin, Bosma, Mishra,
  Roberts, Barham, Chung, Sutton, Gehrmann, Schuh, Shi, Tsvyashchenko, Maynez,
  Rao, Barnes, Tay, Shazeer, Prabhakaran, Reif, Du, Hutchinson, Pope, Bradbury,
  Austin, Isard, {Gur-Ari}, Yin, Duke, Levskaya, Ghemawat, Dev, Michalewski,
  Garcia, Misra, Robinson, Fedus, Zhou, Ippolito, Luan, Lim, Zoph, Spiridonov,
  Sepassi, Dohan, Agrawal, Omernick, Dai, Pillai, Pellat, Lewkowycz, Moreira,
  Child, Polozov, Lee, Zhou, Wang, Saeta, Diaz, Firat, Catasta, Wei,
  {Meier-Hellstern}, Eck, Dean, Petrov, and Fiedel]{chowdhery2023}
Aakanksha Chowdhery, Sharan Narang, Jacob Devlin, Maarten Bosma, Gaurav Mishra,
  Adam Roberts, Paul Barham, Hyung~Won Chung, Charles Sutton, Sebastian
  Gehrmann, Parker Schuh, Kensen Shi, Sasha Tsvyashchenko, Joshua Maynez,
  Abhishek Rao, Parker Barnes, Yi~Tay, Noam Shazeer, Vinodkumar Prabhakaran,
  Emily Reif, Nan Du, Ben Hutchinson, Reiner Pope, James Bradbury, Jacob
  Austin, Michael Isard, Guy {Gur-Ari}, Pengcheng Yin, Toju Duke, Anselm
  Levskaya, Sanjay Ghemawat, Sunipa Dev, Henryk Michalewski, Xavier Garcia,
  Vedant Misra, Kevin Robinson, Liam Fedus, Denny Zhou, Daphne Ippolito, David
  Luan, Hyeontaek Lim, Barret Zoph, Alexander Spiridonov, Ryan Sepassi, David
  Dohan, Shivani Agrawal, Mark Omernick, Andrew~M. Dai,
  Thanumalayan~Sankaranarayana Pillai, Marie Pellat, Aitor Lewkowycz, Erica
  Moreira, Rewon Child, Oleksandr Polozov, Katherine Lee, Zongwei Zhou, Xuezhi
  Wang, Brennan Saeta, Mark Diaz, Orhan Firat, Michele Catasta, Jason Wei,
  Kathy {Meier-Hellstern}, Douglas Eck, Jeff Dean, Slav Petrov, and Noah
  Fiedel.
\newblock {{PaLM}}: {{Scaling}} {L}anguage {M}odeling with {P}athways.
\newblock \emph{Journal of Machine Learning Research}, 24\penalty0
  (240):\penalty0 1--113, 2023.

\bibitem[Dale(2021)]{dale2021}
Robert Dale.
\newblock {{GPT-3}}: {{What}}'s {I}t {G}ood for?
\newblock \emph{Natural Language Engineering}, 27\penalty0 (1):\penalty0
  113--118, 2021.

\bibitem[Haider(2023)]{haider2023}
Hubert Haider.
\newblock Is {{Chat-GPT}} a {G}rammatically {C}ompetent {I}nformant?
\newblock Preprint, lingbuzz 007285, 2023.

\bibitem[Bender and Koller(2020)]{bender2020}
Emily~M. Bender and Alexander Koller.
\newblock ``{C}limbing towards {{NLU}}: {{On Meaning}}, {{Form}}, and
  {{Understanding}} in the {{Age}} of {{Data}}''.
\newblock In \emph{Proceedings of the 58th {{Annual Meeting}} of the
  {{Association}} for {{Computational Linguistics}}}, pages 5185--5198, 2020.

\bibitem[Dentella et~al.(2023{\natexlab{a}})Dentella, Murphy, Marcus, and
  Leivada]{dentella2023}
Vittoria Dentella, Elliot Murphy, Gary Marcus, and Evelina Leivada.
\newblock Testing {{AI}} {P}erformance on {L}ess {F}requent {A}spects of
  {L}anguage {R}eveals {I}nsensitivity to {U}nderlying {M}eaning.
\newblock Preprint, arXiv 2302.12313, 2023{\natexlab{a}}.

\bibitem[Katzir(2023)]{katzir2023}
Roni Katzir.
\newblock Why {L}arge {L}anguage {M}odels {A}re {P}oor {T}heories of {H}uman
  {L}inguistic {C}ognition: {{A}} {R}eply to {{Piantadosi}} (2023).
\newblock Preprint, lingbuzz 007190, 2023.

\bibitem[Weissweiler et~al.(2023)Weissweiler, Hofmann, Kantharuban, Cai, Dutt,
  Hengle, Kabra, Kulkarni, Vijayakumar, Yu, Schütze, Oflazer, and
  Mortensen]{weissweiler2023a}
Leonie Weissweiler, Valentin Hofmann, Anjali Kantharuban, Anna Cai, Ritam Dutt,
  Amey Hengle, Anubha Kabra, Atharva Kulkarni, Abhishek Vijayakumar, Haofei Yu,
  Hinrich Schütze, Kemal Oflazer, and David Mortensen.
\newblock ``{C}ounting the {Bugs} in {C}hat{GPT}{'}s {Wugs}: A {M}ultilingual
  {I}nvestigation into the {M}orphological {C}apabilities of a {L}arge
  {L}anguage {M}odel''.
\newblock In \emph{Proceedings of the 2023 Conference on Empirical Methods in
  Natural Language Processing}, pages 6508--6524, 2023.

\bibitem[Gulordava et~al.(2018)Gulordava, Bojanowski, Grave, Linzen, and
  Baroni]{gulordava2018}
Kristina Gulordava, Piotr Bojanowski, Edouard Grave, Tal Linzen, and Marco
  Baroni.
\newblock ``{C}olorless {{Green Recurrent Networks Dream Hierarchically}}''.
\newblock In \emph{Proceedings of the 2018 {{Conference}} of the {{North
  American Chapter}} of the {{Association}} for {{Computational Linguistics}}:
  {{Human Language Technologies}}}, pages 1195--1205, 2018.

\bibitem[Haley(2020)]{haley2020}
Coleman Haley.
\newblock ``{T}his is a {{BERT}}. {{Now}} {T}here {A}re {S}everal of {T}hem.
  {{Can}} {T}hey {G}eneralize to {N}ovel {W}ords?''.
\newblock In \emph{Proceedings of the {{Third BlackboxNLP Workshop}} on
  {{Analyzing}} and {{Interpreting Neural Networks}} for {{NLP}}}, pages
  333--341, 2020.

\bibitem[Kim and Smolensky(2021)]{kim2021}
Najoung Kim and Paul Smolensky.
\newblock ``{T}esting for {{Grammatical Category Abstraction}} in {{Neural
  Language Models}}''.
\newblock In \emph{Proceedings of the {{Society}} for {{Computation}} in
  {{Linguistics}} 2021}, pages 467--470, 2021.

\bibitem[Maudslay and Cotterell(2021)]{maudslay2021}
Rowan~Hall Maudslay and Ryan Cotterell.
\newblock ``{D}o {{Syntactic Probes Probe Syntax}}? {{Experiments}} with
  {{Jabberwocky Probing}}''.
\newblock In \emph{Proceedings of the 2021 {{Conference}} of the {{North
  American Chapter}} of the {{Association}} for {{Computational Linguistics}}:
  {{Human Language Technologies}}}, pages 124--131, 2021.

\bibitem[Wei et~al.(2021)Wei, Garrette, Linzen, and Pavlick]{wei2021}
Jason Wei, Dan Garrette, Tal Linzen, and Ellie Pavlick.
\newblock ``{F}requency {{Effects}} on {{Syntactic Rule Learning}} in
  {{Transformers}}''.
\newblock In \emph{Proceedings of the 2021 {{Conference}} on {{Empirical
  Methods}} in {{Natural Language Processing}}}, pages 932--948, 2021.

\bibitem[McCoy et~al.(2023)McCoy, Smolensky, Linzen, Gao, and
  Celikyilmaz]{mccoy2023}
R.~Thomas McCoy, Paul Smolensky, Tal Linzen, Jianfeng Gao, and Asli
  Celikyilmaz.
\newblock How {{Much Do Language Models Copy From Their Training Data}}?
  {{Evaluating Linguistic Novelty}} in {{Text Generation Using RAVEN}}.
\newblock \emph{Transactions of the Association for Computational Linguistics},
  11:\penalty0 652--670, 2023.

\bibitem[Chomsky(1965)]{chomsky1965}
Noam Chomsky.
\newblock \emph{Aspects of the {T}heory of {S}yntax}.
\newblock MIT Press, Cambridge, MA, 1965.

\bibitem[Chomsky and Halle(1968)]{chomsky1968}
Noam Chomsky and Morris Halle.
\newblock \emph{The {S}ound {P}attern of {{English}}}.
\newblock Harper \& Row, New York, NY, 1968.

\bibitem[Nosofsky(1986)]{nosofsky1986}
Robert~M. Nosofsky.
\newblock Attention, {{Similarity}}, and the {{Identification-Categorization
  Relationship}}.
\newblock \emph{Journal of Experimental Psychology: General}, 115\penalty0
  (1):\penalty0 39--57, 1986.

\bibitem[Nosofsky(1988)]{nosofsky1988}
Robert~M. Nosofsky.
\newblock Similarity, {{Frequency}}, and {{Category Representations}}.
\newblock \emph{Journal of Experimental Psychology: Learning, Memory, and
  Cognition}, 14\penalty0 (1):\penalty0 54--65, 1988.

\bibitem[Gentner et~al.(2001)Gentner, Holyoak, and Kokinov]{gentner2001}
Dedre Gentner, Keith Holyoak, and Boicho Kokinov.
\newblock \emph{The {A}nalogical {M}ind: {{Perspectives}} from {C}ognitive
  {S}cience}.
\newblock MIT Press, Cambridge, MA, 2001.

\bibitem[Tenenbaum and Griffiths(2001)]{tenenbaum2001}
Joshua~B. Tenenbaum and Thomas~L. Griffiths.
\newblock Generalization, {S}imilarity, and {{Bayesian}} {I}nference.
\newblock \emph{Behavioral and Brain Sciences}, 24\penalty0 (4):\penalty0
  629--640, 2001.

\bibitem[Skousen(1989)]{skousen1989}
Royal Skousen.
\newblock \emph{Analogical {M}odeling of {L}anguage}.
\newblock Kluwer, Dordrecht, 1989.

\bibitem[Johnson(1997)]{johnson1997FIX}
Keith Johnson.
\newblock ``{S}peech {P}erception {W}ithout {S}peaker {N}ormalization: {A}n
  {E}xemplar {M}odel''.
\newblock In Keith Johnson and John~W. Mullennix, editors, \emph{Talker
  {V}ariability in {S}peech {P}rocessing}, pages 145--165. 1997.

\bibitem[{Dawdy-Hesterberg} and Pierrehumbert(2014)]{dawdy-hesterberg2014}
Lisa {Dawdy-Hesterberg} and Janet Pierrehumbert.
\newblock Learnability and {G}eneralisation of {{Arabic}} {B}roken {P}lural
  {N}ouns.
\newblock \emph{Language, Cognition and Neuroscience}, 29\penalty0
  (10):\penalty0 1268--1282, 2014.

\bibitem[Todd et~al.(2019)Todd, Pierrehumbert, and Hay]{todd2019}
Simon Todd, Janet Pierrehumbert, and Jennifer Hay.
\newblock Word {F}requency {E}ffects in {S}ound {C}hange as a {C}onsequence of
  {P}erceptual {A}symmetries: {{An}} {E}xemplar-{B}ased {M}odel.
\newblock \emph{Cognition}, 185:\penalty0 1--20, 2019.

\bibitem[Ambridge(2020)]{ambridge2020}
Ben Ambridge.
\newblock Against {S}tored {A}bstractions: {{A}} {R}adical {E}xemplar {M}odel
  of {L}anguage {A}cquisition.
\newblock \emph{First Language}, 40\penalty0 (5-6):\penalty0 509--559, 2020.

\bibitem[R{\'a}cz et~al.(2020)R{\'a}cz, Beckner, Hay, and
  Pierrehumbert]{racz2020}
P{\'e}ter R{\'a}cz, Clay Beckner, Jennifer~B. Hay, and Janet Pierrehumbert.
\newblock Morphological {C}onvergence as {O}n-{L}ine {L}exical {A}nalogy.
\newblock \emph{Language}, 96\penalty0 (4):\penalty0 735--770, 2020.

\bibitem[R{\'a}cz and Luk{\'a}cs(2024)]{racz2024}
P{\'e}ter R{\'a}cz and {\'A}gnes Luk{\'a}cs.
\newblock Lexical and {S}ocial {E}ffects on the {L}earning and {I}ntegration of
  {I}nflectional {M}orphology.
\newblock \emph{Cognitive Science}, 48\penalty0 (8):\penalty0 e13483, 2024.

\bibitem[Biderman et~al.(2023)Biderman, Prashanth, Sutawika, Schoelkopf,
  Anthony, Purohit, and Raf]{biderman2023}
Stella Biderman, USVSN~Sai Prashanth, Lintang Sutawika, Hailey Schoelkopf,
  Quentin Anthony, Shivanshu Purohit, and Edward Raf.
\newblock Emergent and {{Predictable Memorization}} in {{Large Language
  Models}}.
\newblock Preprint, arXiv 2304.11158, 2023.

\bibitem[Cao et~al.(2023)Cao, Tang, Lin, Han, Chen, Wang, and Sun]{cao2023}
Boxi Cao, Qiaoyu Tang, Hongyu Lin, Xianpei Han, Jiawei Chen, Tianshu Wang, and
  Le~Sun.
\newblock Retentive or {{Forgetful}}? {{Diving}} into the {{Knowledge
  Memorizing Mechanism}} of {{Language Models}}.
\newblock Preprint, arXiv 2305.09144, 2023.

\bibitem[Carlini et~al.(2023)Carlini, Ippolito, Jagielski, Lee, Tramer, and
  Zhang]{carlini2023}
Nicholas Carlini, Daphne Ippolito, Matthew Jagielski, Katherine Lee, Florian
  Tramer, and Chiyuan Zhang.
\newblock ``{Q}uantifying {{Memorization Across Neural Language Models}}''.
\newblock In \emph{International {{Conference}} on {{Learning Representations}}
  2023}, 2023.

\bibitem[Pinker and Prince(1988)]{pinker1988}
Steven Pinker and Alan Prince.
\newblock On {L}anguage and {C}onnectionism: {{Analysis}} of a {P}arallel
  {D}istributed {P}rocessing {M}odel of {L}anguage {A}cquisition.
\newblock \emph{Cognition}, 28\penalty0 (1-2):\penalty0 73--193, 1988.

\bibitem[Pinker and Prince(1991)]{pinker1991FIX}
Steven Pinker and Alan Prince.
\newblock ``{R}egular and {I}rregular {M}orphology and the {P}sychological
  {S}tatus of {R}ules of {G}rammar''.
\newblock In \emph{Annual Meeting of the Berkeley Linguistics Society}, pages
  230--251, 1991.

\bibitem[Prasada and Pinker(1993)]{prasada1993}
Sandeep Prasada and Steven Pinker.
\newblock Generalisation of {R}egular and {I}rregular {M}orphological
  {P}atterns.
\newblock \emph{Language and Cognitive Processes}, 8\penalty0 (1):\penalty0
  1--56, 1993.

\bibitem[Hahn and Chater(1998)]{hahn1998}
Ulrike Hahn and Nick Chater.
\newblock Similarity and {R}ules: {D}istinct? {E}xhaustive? {E}mpirically
  {D}istinguishable?
\newblock \emph{Cognition}, 65\penalty0 (2-3):\penalty0 197--230, 1998.

\bibitem[Pothos(2005)]{pothos2005}
Emmanuel~M. Pothos.
\newblock The {R}ules versus {S}imilarity {D}istinction.
\newblock \emph{Behavioral and Brain Sciences}, 28\penalty0 (1):\penalty0
  1--14, 2005.

\bibitem[{Arndt-Lappe}(2014{\natexlab{a}})]{arndt-lappe2014a}
Sabine {Arndt-Lappe}.
\newblock ``{W}ord-{F}ormation and {A}nalogy''.
\newblock In Peter~O. Müller, Ingeborg Ohnheiser, Susan Olsen, and Franz
  Rainer, editors, \emph{Word-{{Formation}}: {{An International Handbook}} of
  the {{Languages}} of {{Europe}}}, pages 822--841. 2014{\natexlab{a}}.

\bibitem[Rumelhart and McClelland(1986)]{rumelhart1985}
David~E. Rumelhart and James~L. McClelland.
\newblock On {L}earning the {P}ast {T}enses of {E}nglish {V}erbs.
\newblock \emph{Psycholinguistics: Critical Concepts in Psychology},
  4:\penalty0 216--271, 1986.

\bibitem[Plaut and McClelland(1993)]{plaut1993FIX}
David~C Plaut and James~L McClelland.
\newblock ``{G}eneralization {W}ith {C}omponential {A}ttractors: {W}ord and
  {N}onword {R}eading in an {A}ttractor {N}etwork'''.
\newblock In \emph{Proceedings of the 15th Annual Conference of the Cognitive
  Science Society}, pages 824--829, 1993.

\bibitem[Rumelhart and Todd(1993)]{rumelhart1993FIX}
David~E Rumelhart and Peter~M Todd.
\newblock ``{L}earning and {C}onnectionist {R}epresentations''.
\newblock In David~E. Meyer and Sylvan Kornblum, editors, \emph{Attention and
  Performance XIV: Synergies in Experimental Psychology, Artificial
  Intelligence, and Cognitive Neuroscience}, pages 3--30. 1993.

\bibitem[Plaut et~al.(1996)Plaut, McClelland, Seidenberg, and
  Patterson]{plaut1996}
David~C Plaut, James McClelland, Mark~S. Seidenberg, and Karalyn Patterson.
\newblock Understanding {{Normal}} and {{Impaired Word Reading}}:
  {{Computational Principles}} in {{Quasi-Regular Domains}}.
\newblock \emph{Psychological Review}, 103\penalty0 (1):\penalty0 56--115,
  1996.

\bibitem[Plaut and Gonnerman(2000)]{plaut2000}
David~C. Plaut and Laura~M. Gonnerman.
\newblock Are {N}on-{S}emantic {M}orphological {E}ffects {I}ncompatible with a
  {D}istributed {C}onnectionist {A}pproach to {L}exical {P}rocessing?
\newblock \emph{Language and Cognitive Processes}, 15\penalty0 (4-5):\penalty0
  445--485, 2000.

\bibitem[Gonnerman et~al.(2007)Gonnerman, Seidenberg, and
  Andersen]{Gonnerman.2007}
Laura~M. Gonnerman, Mark~S. Seidenberg, and Elaine~S. Andersen.
\newblock Graded {S}emantic and {P}honological {S}imilarity {E}ffects in
  {P}riming: {{Evidence}} for a {D}istributed {C}onnectionist {A}pproach to
  {M}orphology.
\newblock \emph{Journal of Experimental Psychology: General}, 136\penalty0
  (2):\penalty0 323--345, 2007.

\bibitem[Aronoff(1976)]{aronoff1976}
Mark Aronoff.
\newblock \emph{Word {{Formation}} in {{Generative Grammar}}}.
\newblock MIT Press, Cambridge, MA, 1976.

\bibitem[Bauer(2001)]{Bauer.2001}
Laurie Bauer.
\newblock \emph{Morphological {P}roductivity}.
\newblock Cambridge University Press, Cambridge, UK, 2001.

\bibitem[Haspelmath and Sims(2010)]{haspelmath2010}
Martin Haspelmath and Andrea Sims.
\newblock \emph{Understanding {M}orphology}.
\newblock Routledge, Oxford, UK, 2010.

\bibitem[Bauer et~al.(2013)Bauer, Lieber, and Plag]{Bauer.2013}
Laurie Bauer, Rochelle Lieber, and Ingo Plag.
\newblock \emph{The {{Oxford}} {R}eference {G}uide to {{English}}
  {M}orphology}.
\newblock Oxford University Press, Oxford, UK, 2013.

\bibitem[Beard(2017)]{beard2017derivation}
Robert Beard.
\newblock ``{D}erivation''.
\newblock In Andrew Spencer and Arnold~M. Zwicky, editors, \emph{The {H}andbook
  of {M}orphology}, pages 44--65, 2017.

\bibitem[Anshen and Aronoff(1988)]{anshen1988FIX}
Frank Anshen and Mark Aronoff.
\newblock Producing {M}orphologically {C}omplex {W}ords.
\newblock \emph{Linguistics}, 26\penalty0 (4):\penalty0 641--656, 1988.

\bibitem[Baayen and Renouf(1996)]{baayen1996}
R.~Harald Baayen and Antoinette Renouf.
\newblock Chronicling the {{Times}}: {{Productive {L}exical {I}nnovations}} in
  an {{English {N}ewspaper}}.
\newblock \emph{Language}, 72\penalty0 (1):\penalty0 69--96, 1996.

\bibitem[Anshen and Aronoff(1999)]{anshen1999}
Frank Anshen and Mark Aronoff.
\newblock Using {{Dictionaries}} to {{Study}} the {{Mental Lexicon}}.
\newblock \emph{Brain and Language}, 68\penalty0 (1-2):\penalty0 16--26, 1999.

\bibitem[Lindsay(2012)]{lindsay2012}
Mark Lindsay.
\newblock Rival {S}uffixes: {S}ynonymy, {C}ompetition, and the {E}mergence of
  {P}roductivity.
\newblock \emph{Mediterranean Morphology Meetings}, 8\penalty0 (0):\penalty0
  192--203, 2012.

\bibitem[{Arndt-Lappe}(2014{\natexlab{b}})]{arndt-lappe2014}
Sabine {Arndt-Lappe}.
\newblock Analogy in {S}uffix {R}ivalry: {{The}} {C}ase of {{English}} -{I}ty
  and -{N}ess.
\newblock \emph{English Language \& Linguistics}, 18\penalty0 (3):\penalty0
  497--548, 2014{\natexlab{b}}.

\bibitem[Bresnan et~al.(2007)Bresnan, Cueni, Nikitina, and
  Baayen]{bresnan2007predicting}
Joan Bresnan, Anna Cueni, Tatiana Nikitina, and R~Harald Baayen.
\newblock ``{P}redicting the {D}ative {A}lternation''.
\newblock In Gerlof Bouma, Irene Krämer, and Joost Zwarts, editors,
  \emph{Cognitive Foundations of Interpretation}, pages 69--94. 2007.

\bibitem[Walsh et~al.(2010)Walsh, M{\"o}bius, Wade, and Sch{\"u}tze]{walsh2010}
Michael Walsh, Bernd M{\"o}bius, Travis Wade, and Hinrich Sch{\"u}tze.
\newblock Multilevel {{Exemplar Theory}}.
\newblock \emph{Cognitive Science}, 34\penalty0 (4):\penalty0 537--582, 2010.

\bibitem[Bresnan(2021)]{bresnan2021}
Joan Bresnan.
\newblock Formal {G}rammar, {U}sage {P}robabilities, and {A}uxiliary
  {C}ontraction.
\newblock \emph{Language}, 97\penalty0 (1):\penalty0 108--150, 2021.

\bibitem[Edmiston(2020)]{edmiston2020}
Daniel Edmiston.
\newblock A {{Systematic Analysis}} of {{Morphological Content}} in {{BERT
  Models}} for {{Multiple Languages}}.
\newblock Preprint, arXiv 2004.03032, 2020.

\bibitem[Hofmann et~al.(2020{\natexlab{a}})Hofmann, Pierrehumbert, and
  Sch{\"u}tze]{hofmann2020}
Valentin Hofmann, Janet Pierrehumbert, and Hinrich Sch{\"u}tze.
\newblock ``{{DagoBERT}}: {{Generating Derivational Morphology}} with a
  {{Pretrained Language Model}}''.
\newblock In \emph{Proceedings of the 2020 {{Conference}} on {{Empirical
  Methods}} in {{Natural Language Processing}}}, pages 3848--3861,
  2020{\natexlab{a}}.

\bibitem[Hofmann et~al.(2021)Hofmann, Pierrehumbert, and
  Sch{\"u}tze]{hofmann2021}
Valentin Hofmann, Janet Pierrehumbert, and Hinrich Sch{\"u}tze.
\newblock ``{S}uperbizarre {{Is Not Superb}}: {{Derivational Morphology
  Improves BERT}}'s {{Interpretation}} of {{Complex Words}}''.
\newblock In \emph{Proceedings of the 59th {{Annual Meeting}} of the
  {{Association}} for {{Computational Linguistics}}}, pages 3594--3608, 2021.

\bibitem[Sun(2008)]{sun2008}
Ron Sun.
\newblock ``{I}ntroduction to {C}omputational {C}ognitive {M}odeling''.
\newblock In Ron Sun, editor, \emph{The {{Cambridge}} Handbook of Computational
  Psychology}, pages 3--19. 2008.

\bibitem[Wilson and Collins(2019)]{wilson2019}
Robert Wilson and Anne Collins.
\newblock Ten {S}imple {R}ules for the {C}omputational {M}odeling of
  {B}ehavioral {D}ata.
\newblock \emph{eLife}, 8:\penalty0 e49547, 2019.

\bibitem[Brasoveanu and Dotla{\v c}il(2020)]{brasoveanu2020}
Adrian Brasoveanu and Jakub Dotla{\v c}il.
\newblock \emph{Computational {{Cognitive Modeling}} and {{Linguistic
  Theory}}}.
\newblock Springer, Cham, 2020.

\bibitem[Akyurek et~al.(2022)Akyurek, Bolukbasi, Liu, Xiong, Tenney, Andreas,
  and Guu]{akyurek2022b}
Ekin Akyurek, Tolga Bolukbasi, Frederick Liu, Binbin Xiong, Ian Tenney, Jacob
  Andreas, and Kelvin Guu.
\newblock ``{T}owards {{Tracing Knowledge}} in {{Language Models Back}} to the
  {{Training Data}}''.
\newblock In \emph{Findings of the {{Association}} for {{Computational
  Linguistics}}: {{EMNLP}} 2022}, pages 2429--2446, 2022.

\bibitem[Han and Tsvetkov(2022)]{han2022}
Xiaochuang Han and Yulia Tsvetkov.
\newblock {{ORCA}}: {{Interpreting Prompted Language Models}} via {{Locating
  Supporting Data Evidence}} in the {{Ocean}} of {{Pretraining Data}}.
\newblock Preprint, arXiv 2205.12600, 2022.

\bibitem[Razeghi et~al.(2022)Razeghi, Logan~IV, Gardner, and
  Singh]{razeghi2022}
Yasaman Razeghi, Robert Logan~IV, Matt Gardner, and Sameer Singh.
\newblock ``{I}mpact of {{Pretraining Term Frequencies}} on {{Few-Shot
  Numerical Reasoning}}''.
\newblock In \emph{Findings of the {{Association}} for {{Computational
  Linguistics}}: {{EMNLP}} 2022}, pages 840--854, 2022.

\bibitem[Elazar et~al.(2023)Elazar, Kassner, Ravfogel, Feder, Ravichander,
  Mosbach, Belinkov, Sch{\"u}tze, and Goldberg]{elazar2023}
Yanai Elazar, Nora Kassner, Shauli Ravfogel, Amir Feder, Abhilasha Ravichander,
  Marius Mosbach, Yonatan Belinkov, Hinrich Sch{\"u}tze, and Yoav Goldberg.
\newblock Measuring {{Causal Effects}} of {{Data Statistics}} on {{Language
  Model}}'s `{{Factual}}' {{Predictions}}.
\newblock Preprint, arXiv 2207.14251, 2023.

\bibitem[Wang and Komatsuzaki(2021)]{wang2021a}
Ben Wang and Aran Komatsuzaki.
\newblock {{GPT-J-6B}}: {{A}} 6 {{Billion Parameter Autoregressive Language
  Model}}.
\newblock https://github.com/kingoflolz/mesh-transformer-jax, 2021.

\bibitem[Gao et~al.(2020)Gao, Biderman, Black, Golding, Hoppe, Foster, Phang,
  He, Thite, Nabeshima, Presser, and Leahy]{gao2020}
Leo Gao, Stella Biderman, Sid Black, Laurence Golding, Travis Hoppe, Charles
  Foster, Jason Phang, Horace He, Anish Thite, Noa Nabeshima, Shawn Presser,
  and Connor Leahy.
\newblock The {P}ile: {A}n 800{GB} {D}ataset of {D}iverse {T}ext for {L}anguage
  {M}odeling.
\newblock Preprint, arXiv 2101.00027, 2020.

\bibitem[Albright and Hayes(2002)]{albright2002}
Adam Albright and Bruce Hayes.
\newblock ``{M}odeling {{English Past Tense Intuitions}} with {{Minimal
  Generalization}}''.
\newblock In \emph{Proceedings of the {{ACL-02 Workshop}} on {{Morphological}}
  and {{Phonological Learning}}}, pages 58--69, 2002.

\bibitem[Albright and Hayes(2003)]{albright2003FIX}
Adam Albright and Bruce Hayes.
\newblock Rules vs. {A}nalogy in {{English}} {P}ast {T}enses: {A}
  {C}omputational/{E}xperimental {S}tudy.
\newblock \emph{Cognition}, 90\penalty0 (2):\penalty0 119--161, 2003.

\bibitem[Nosofsky(1990)]{nosofsky1990}
Robert~M. Nosofsky.
\newblock Relations between {E}xemplar-{S}imilarity and {L}ikelihood {M}odels
  of {C}lassification.
\newblock \emph{Journal of Mathematical Psychology}, 34\penalty0 (4):\penalty0
  393--418, 1990.

\bibitem[New et~al.(2024)New, Bourgin, Barra, and Pallier]{new2023FIX}
Boris New, Jessica Bourgin, Julien Barra, and Christophe Pallier.
\newblock Uni{P}seudo: {A} {U}niversal {P}seudoword {G}enerator.
\newblock \emph{Quarterly Journal of Experimental Psychology}, 77\penalty0
  (2):\penalty0 278--286, 2024.

\bibitem[Rae et~al.(2022)Rae, Borgeaud, Cai, Millican, Hoffmann, Song,
  Aslanides, Henderson, Ring, Young, Rutherford, Hennigan, Menick, Cassirer,
  Powell, van~den Driessche, Hendricks, Rauh, Huang, Glaese, Welbl, Dathathri,
  Huang, Uesato, Mellor, Higgins, Creswell, McAleese, Wu, Elsen, Jayakumar,
  Buchatskaya, Budden, Sutherland, Simonyan, Paganini, Sifre, Martens, Li,
  Kuncoro, Nematzadeh, Gribovskaya, Donato, Lazaridou, Mensch, Lespiau,
  Tsimpoukelli, Grigorev, Fritz, Sottiaux, Pajarskas, Pohlen, Gong, Toyama,
  de~Masson~d'Autume, Li, Terzi, Mikulik, Babuschkin, Clark, de~Las~Casas, Guy,
  Jones, Bradbury, Johnson, Hechtman, Weidinger, Gabriel, Isaac, Lockhart,
  Osindero, Rimell, Dyer, Vinyals, Ayoub, Stanway, Bennett, Hassabis,
  Kavukcuoglu, and Irving]{rae2022FIX}
Jack~W. Rae, Sebastian Borgeaud, Trevor Cai, Katie Millican, Jordan Hoffmann,
  Francis Song, John Aslanides, Sarah Henderson, Roman Ring, Susannah Young,
  Eliza Rutherford, Tom Hennigan, Jacob Menick, Albin Cassirer, Richard Powell,
  George van~den Driessche, Lisa~Anne Hendricks, Maribeth Rauh, Po-Sen Huang,
  Amelia Glaese, Johannes Welbl, Sumanth Dathathri, Saffron Huang, Jonathan
  Uesato, John Mellor, Irina Higgins, Antonia Creswell, Nat McAleese, Amy Wu,
  Erich Elsen, Siddhant Jayakumar, Elena Buchatskaya, David Budden, Esme
  Sutherland, Karen Simonyan, Michela Paganini, Laurent Sifre, Lena Martens,
  Xiang~Lorraine Li, Adhiguna Kuncoro, Aida Nematzadeh, Elena Gribovskaya,
  Domenic Donato, Angeliki Lazaridou, Arthur Mensch, Jean-Baptiste Lespiau,
  Maria Tsimpoukelli, Nikolai Grigorev, Doug Fritz, Thibault Sottiaux, Mantas
  Pajarskas, Toby Pohlen, Zhitao Gong, Daniel Toyama, Cyprien
  de~Masson~d'Autume, Yujia Li, Tayfun Terzi, Vladimir Mikulik, Igor
  Babuschkin, Aidan Clark, Diego de~Las~Casas, Aurelia Guy, Chris Jones, James
  Bradbury, Matthew Johnson, Blake Hechtman, Laura Weidinger, Iason Gabriel,
  William Isaac, Ed~Lockhart, Simon Osindero, Laura Rimell, Chris Dyer, Oriol
  Vinyals, Kareem Ayoub, Jeff Stanway, Lorrayne Bennett, Demis Hassabis, Koray
  Kavukcuoglu, and Geoffrey Irving.
\newblock Scaling {L}anguage {M}odels: {M}ethods, {A}nalysis \& {I}nsights from
  {T}raining {G}opher.
\newblock Preprint, arXiv 2112.11446, 2022.

\bibitem[Hare et~al.(2001)Hare, Ford, and Marslen-Wilson]{hare2001}
Mary~L. Hare, Marilyn Ford, and William~D. Marslen-Wilson.
\newblock ``{A}mbiguity and {F}requency {E}ffects in {R}egular {V}erb
  {I}nflection''.
\newblock In Joan Bybee and Paul Hopper, editors, \emph{Frequency and the
  {E}mergence of {L}inguistic {S}tructure}, pages 181--200. 2001.

\bibitem[Arndt-Lappe and Ernestus(2020)]{arndt-lappe2020}
Sabine Arndt-Lappe and Mirjam Ernestus.
\newblock ``{M}orpho-{P}honological {A}lternations: {T}he {R}ole of {L}exical
  {S}torage''.
\newblock In V.~Pirelli, I.~Plag, and W.U. Dressler, editors, \emph{Word
  Knowledge and Word Usage: A Cross-Disciplinary Guide to the Mental Lexicon},
  pages 191--227. 2020.

\bibitem[Holm(1979)]{Holm.1979}
Sture Holm.
\newblock A {S}imple {S}equentially {R}ejective {M}ultiple {T}est {P}rocedure.
\newblock \emph{Scandinavian Journal of Statistics}, 6\penalty0 (2):\penalty0
  65--70, 1979.

\bibitem[Lakoff(1970)]{lakoff1970}
George Lakoff.
\newblock \emph{Irregularity in {S}yntax}.
\newblock Holt, Rinehart and Winston, New York City, NY, 1970.

\bibitem[Corbett et~al.(2001)Corbett, Hippisley, Brown, and
  Marriott]{corbett2001}
Greville Corbett, Andrew Hippisley, Dunstan Brown, and Paul Marriott.
\newblock ``{F}requency, {R}egularity, and the {P}aradigm''.
\newblock In Joan Bybee and Paul Hopper, editors, \emph{Frequency and the
  {E}mergence of {L}inguistic {S}tructure}, pages 201--228. 2001.

\bibitem[Bybee(2006)]{bybee2006}
Joan Bybee.
\newblock From {U}sage to {G}rammar: {T}he {M}ind’s {R}esponse to
  {R}epetition.
\newblock \emph{Language}, 82\penalty0 (4):\penalty0 711--733, 2006.

\bibitem[Lieberman et~al.(2007)Lieberman, Michel, Jackson, Tang, and
  Nowak]{lieberman2007}
Erez Lieberman, Jean-Baptiste Michel, Joe Jackson, Tina Tang, and Martin~A.
  Nowak.
\newblock Quantifying the {R}evolutionary {D}ynamics of {L}anguage.
\newblock \emph{Nature}, 449\penalty0 (7163):\penalty0 713--716, 2007.

\bibitem[Daland et~al.(2007)Daland, Sims, and Pierrehumbert]{daland2007new}
Robert Daland, Andrea~D. Sims, and Janet Pierrehumbert.
\newblock ``{M}uch {A}do about {N}othing: {A} {S}ocial {N}etwork {M}odel of
  {R}ussian {P}aradigmatic {G}aps''.
\newblock In \emph{Proceedings of the 45th Annual Meeting of the Association of
  Computational Linguistics}, pages 936--943, 2007.

\bibitem[Petrov et~al.(2024)Petrov, Torr, and Bibi]{petrov2024}
Aleksandar Petrov, Philip~H.S. Torr, and Adel Bibi.
\newblock ``{W}hen do {P}rompting and {P}refix-{T}uning {W}ork? {A} {T}heory of
  {C}apabilities and {L}imitations''.
\newblock In \emph{International Conference on Learning Representations 2024},
  2024.

\bibitem[Bybee(1995)]{bybee1995}
Joan Bybee.
\newblock Regular {M}orphology and the {L}exicon.
\newblock \emph{Language and Cognitive Processes}, 10\penalty0 (5):\penalty0
  425--455, 1995.

\bibitem[Pierrehumbert(2001)]{pierrehumbert2001}
Janet Pierrehumbert.
\newblock Stochastic {P}honology.
\newblock \emph{Glot International}, 5\penalty0 (6):\penalty0 195--207, 2001.

\bibitem[Pierrehumbert(2003)]{pierrehumbert2003}
Janet Pierrehumbert.
\newblock Phonetic {D}iversity, {S}tatistical {L}earning, and the {A}cquisition
  of {P}honology.
\newblock \emph{Language and Speech}, 46\penalty0 (2-3):\penalty0 115--154,
  2003.

\bibitem[McKenzie et~al.(2023)McKenzie, Lyzhov, Pieler, Parrish, Mueller,
  Prabhu, McLean, Kirtland, Ross, Liu, Gritsevskiy, Wurgaft, Kauffman, Recchia,
  Liu, Cavanagh, Weiss, Huang, Droid, Tseng, Korbak, Shen, Zhang, Zhou, Kim,
  Bowman, and Perez]{mckenzie2023}
Ian~R. McKenzie, Alexander Lyzhov, Michael Pieler, Alicia Parrish, Aaron
  Mueller, Ameya Prabhu, Euan McLean, Aaron Kirtland, Alexis Ross, Alisa Liu,
  Andrew Gritsevskiy, Daniel Wurgaft, Derik Kauffman, Gabriel Recchia, Jiacheng
  Liu, Joe Cavanagh, Max Weiss, Sicong Huang, The~Floating Droid, Tom Tseng,
  Tomasz Korbak, Xudong Shen, Yuhui Zhang, Zhengping Zhou, Najoung Kim,
  Samuel~R. Bowman, and Ethan Perez.
\newblock Inverse {{Scaling}}: {{When Bigger Isn}}'t {{Better}}.
\newblock Preprint, arXiv 2306.09479, 2023.

\bibitem[Stump(2017)]{Stump.2017}
Gregory Stump.
\newblock Rule {C}onflation in an {I}nferential-{R}ealizational {T}heory of
  {M}orphotactics.
\newblock \emph{Acta Linguistica Academica}, 64\penalty0 (1):\penalty0 79--124,
  2017.

\bibitem[Stump(2019)]{Stump.2019}
Gregory Stump.
\newblock Some {S}ources of {A}pparent {G}aps in {D}erivational {P}aradigms.
\newblock \emph{Morphology}, 29\penalty0 (2):\penalty0 271--292, 2019.

\bibitem[Pierrehumbert(2006)]{Pierrehumbert.2006FIX}
Janet Pierrehumbert.
\newblock The {S}tatistical {B}asis of an {U}nnatural {A}lternation,.
\newblock In Louis Goldstein, Douglas~H. Whalen, and Catherine~T. Best,
  editors, \emph{Laboratory Phonology 8}, pages 81--106. 2006.

\bibitem[Säily(2011)]{saily2011FIX}
Tanja Säily.
\newblock Variation in {M}orphological {P}roductivity in the {BNC}:
  {S}ociolinguistic and {M}ethodological {C}onsiderations.
\newblock \emph{Corpus Linguistics and Linguistic Theory}, 7\penalty0
  (1):\penalty0 119--141, 2011.

\bibitem[S{\"a}ily(2014)]{saily2014}
Tanja S{\"a}ily.
\newblock \emph{Sociolinguistic {{Variation}} in {{English Derivational
  Productivity}}: {{Studies}} and {{Methods}} in {{Diachronic Corpus
  Linguistics}}}.
\newblock Soci{\'e}t{\'e} N{\'e}ophilologique de Helsinki, Helsinki, 2014.

\bibitem[S{\"a}ily(2016)]{saily2016}
Tanja S{\"a}ily.
\newblock Sociolinguistic {V}ariation in {M}orphological {P}roductivity in
  {E}ighteenth-{C}entury {{English}}.
\newblock \emph{Corpus Linguistics and Linguistic Theory}, 12\penalty0
  (1):\penalty0 129--151, 2016.

\bibitem[Pierrehumbert(2016)]{pierrehumbert2016FIX}
Janet Pierrehumbert.
\newblock Phonological {R}epresentation: {B}eyond {A}bstract versus {E}pisodic.
\newblock \emph{Annual Review of Linguistics}, 2\penalty0 (1):\penalty0 33--52,
  2016.

\bibitem[Nusbaum et~al.(1984)Nusbaum, Pisoni, and Davis]{nusbaum1984FIX}
Howard~C Nusbaum, David~B Pisoni, and Christopher~K Davis.
\newblock Sizing up the {H}oosier {M}ental {L}exicon.
\newblock \emph{Research on Spoken Language Processing Report}, 10\penalty0
  (3):\penalty0 357--376, 1984.

\bibitem[Needle et~al.(2022)Needle, Pierrehumbert, and Hay]{needle2022}
Jeremy~M. Needle, Janet Pierrehumbert, and Jennifer~B. Hay.
\newblock ``{P}honotactic and {{Morphological Effects}} in the
  {{Acceptability}} of {{Pseudowords}}''.
\newblock In Andrea~D. Sims, Adam Ussishkin, Jeff Parker, and Samantha Wray,
  editors, \emph{Morphological {{Diversity}} and {{Linguistic Cognition}}},
  pages 79--112. 2022.

\bibitem[Hofmann et~al.(2020{\natexlab{b}})Hofmann, Pierrehumbert, and
  Sch{\"u}tze]{hofmann2020a}
Valentin Hofmann, Janet Pierrehumbert, and Hinrich Sch{\"u}tze.
\newblock ``{P}redicting the {{Growth}} of {{Morphological Families}} from
  {{Social}} and {{Linguistic Factors}}''.
\newblock In \emph{Proceedings of the 58th {{Annual Meeting}} of the
  {{Association}} for {{Computational Linguistics}}}, pages 7273--7283,
  2020{\natexlab{b}}.

\bibitem[Hofmann et~al.(2020{\natexlab{c}})Hofmann, Sch{\"u}tze, and
  Pierrehumbert]{hofmann-etal-2020-graph}
Valentin Hofmann, Hinrich Sch{\"u}tze, and Janet Pierrehumbert.
\newblock ``{A} {Graph} {Auto-Encoder} {Model of Derivational Morphology}.
\newblock In \emph{Proceedings of the 58th Annual Meeting of the Association
  for Computational Linguistics}, pages 1127--1138, 2020{\natexlab{c}}.

\bibitem[Wilcox et~al.(2023)Wilcox, Pimentel, Meister, Cotterell, and
  Levy]{wilcox2023}
Ethan~G. Wilcox, Tiago Pimentel, Clara Meister, Ryan Cotterell, and Roger~P.
  Levy.
\newblock Testing the {{Predictions}} of {{Surprisal Theory}} in 11
  {{Languages}}.
\newblock \emph{Transactions of the Association for Computational Linguistics},
  11:\penalty0 1451--1470, 2023.

\bibitem[Brysbaert et~al.(2019)Brysbaert, Mandera, McCormick, and
  Keuleers]{brysbaert2019}
Marc Brysbaert, Pawe{\l} Mandera, Samantha~F. McCormick, and Emmanuel Keuleers.
\newblock Word {P}revalence {N}orms for 62,000 {{English}} {L}emmas.
\newblock \emph{Behavior Research Methods}, 51\penalty0 (2):\penalty0 467--479,
  2019.

\bibitem[Rastle et~al.(2004)Rastle, Davis, and New]{Rastle.2004}
Kathleen Rastle, Matthew~H. Davis, and Boris New.
\newblock The {B}roth in {M}y {B}rother's {B}rothel: {{Morpho-{O}rthographic}}
  {S}egmentation in {V}isual {W}ord {R}ecognition.
\newblock \emph{Psychonomic Bulletin and Review}, 11\penalty0 (6):\penalty0
  1090--1098, 2004.

\bibitem[Taft(2004)]{Taft.2004}
Marcus Taft.
\newblock Morphological {D}ecomposition and the {R}everse {B}ase {F}requency
  {E}ffect.
\newblock \emph{The Quarterly Journal of Experimental Psychology}, 57\penalty0
  (4):\penalty0 745--765, 2004.

\bibitem[Beyersmann et~al.(2015)Beyersmann, Ziegler, and
  Grainger]{Beyersmann.2015}
Elisabeth Beyersmann, Johannes~C. Ziegler, and Jonathan Grainger.
\newblock Differences in the {P}rocessing of {P}refixes and {S}uffixes
  {R}evealed by a {L}etter-{S}earch {T}ask.
\newblock \emph{Scientific Studies of Reading}, 19\penalty0 (5):\penalty0
  360--373, 2015.

\bibitem[Dentella et~al.(2023{\natexlab{b}})Dentella, Günther, and
  Leivada]{dentella}
Vittoria Dentella, Fritz Günther, and Evelina Leivada.
\newblock Systematic {T}esting of {T}hree {L}anguage {M}odels {R}eveals {L}ow
  {L}anguage {A}ccuracy, {A}bsence of {R}esponse {S}tability, and a
  {Y}es-{R}esponse {B}ias.
\newblock \emph{Proceedings of the National Academy of Sciences}, 120\penalty0
  (51):\penalty0 e2309583120, 2023{\natexlab{b}}.

\bibitem[Hu and Levy(2023)]{hu2023prompting}
Jennifer Hu and Roger Levy.
\newblock ``{P}rompting is {N}ot a {S}ubstitute for {P}robability
  {M}easurements in {L}arge {L}anguage {M}odels''.
\newblock In \emph{Proceedings of the 2023 Conference on Empirical Methods in
  Natural Language Processing}, pages 5040--5060, 2023.

\bibitem[Hu et~al.(2024)Hu, Mahowald, Lupyan, Ivanova, and
  Levy]{hu2024language}
Jennifer Hu, Kyle Mahowald, Gary Lupyan, Anna Ivanova, and Roger Levy.
\newblock Language {M}odels {A}lign with {H}uman {J}udgments on {K}ey
  {G}rammatical {C}onstructions.
\newblock \emph{Proceedings of the National Academy of Sciences}, 121\penalty0
  (36):\penalty0 e2400917121, 2024.

\bibitem[Quine(1960)]{quine1960}
Willard~V. Quine.
\newblock \emph{Word and {O}bject}.
\newblock MIT Press, Cambridge, MA, 1960.

\bibitem[Dziri et~al.(2023)Dziri, Lu, Sclar, Li, Jiang, Lin, West, Bhagavatula,
  Bras, Hwang, Sanyal, Welleck, Ren, Ettinger, Harchaoui, and Choi]{dziri2023}
Nouha Dziri, Ximing Lu, Melanie Sclar, Xiang~Lorraine Li, Liwei Jiang,
  Bill~Yuchen Lin, Peter West, Chandra Bhagavatula, Ronan~Le Bras, Jena~D.
  Hwang, Soumya Sanyal, Sean Welleck, Xiang Ren, Allyson Ettinger, Zaid
  Harchaoui, and Yejin Choi.
\newblock Faith and {{Fate}}: {{Limits}} of {{Transformers}} on
  {{Compositionality}}.
\newblock Preprint, arXiv 2305.18654, 2023.

\bibitem[Baayen et~al.(1995)Baayen, Piepenbrock, and Gulikers]{baayen1995}
R.~Harald Baayen, Richard Piepenbrock, and Leon Gulikers.
\newblock \emph{The {{CELEX}} {L}exical {D}atabase ({{CD-ROM}})}.
\newblock Linguistic Data Consortium, Philadelphia, PA, 1995.

\bibitem[{S{\'a}nchez-Guti{\'e}rrez} et~al.(2018){S{\'a}nchez-Guti{\'e}rrez},
  Mailhot, Deacon, and Wilson]{sanchez-gutierrez2018}
Claudia~H. {S{\'a}nchez-Guti{\'e}rrez}, Hugo Mailhot, S.~H{\'e}l{\`e}ne Deacon,
  and Maximiliano~A. Wilson.
\newblock {{MorphoLex}}: {{A}} {D}erivational {M}orphological {D}atabase for
  70,000 {{English}} {W}ords.
\newblock \emph{Behavior Research Methods}, 50\penalty0 (4):\penalty0
  1568--1580, 2018.

\bibitem[Laws and Ryder(2014)]{laws2014}
Jacqueline Laws and Chris Ryder.
\newblock Getting the {M}easure of {D}erivational {M}orphology in {A}dult
  {S}peech: {A} {C}orpus {A}nalysis using {{MorphoQuantics}}.
\newblock \emph{Language Studies Working Papers}, 6:\penalty0 3--17, 2014.

\bibitem[Wolf et~al.(2020)Wolf, Debut, Sanh, Chaumond, Delangue, Moi, Cistac,
  Rault, Louf, Funtowicz, Davison, Shleifer, {von Platen}, Ma, Jernite, Plu,
  Xu, Le~Scao, Gugger, Drame, Lhoest, and Rush]{wolf2020}
Thomas Wolf, Lysandre Debut, Victor Sanh, Julien Chaumond, Clement Delangue,
  Anthony Moi, Pierric Cistac, Tim Rault, Remi Louf, Morgan Funtowicz, Joe
  Davison, Sam Shleifer, Patrick {von Platen}, Clara Ma, Yacine Jernite, Julien
  Plu, Canwen Xu, Teven Le~Scao, Sylvain Gugger, Mariama Drame, Quentin Lhoest,
  and Alexander Rush.
\newblock ``{T}ransformers: {{State-of-the-Art Natural Language Processing}}''.
\newblock In \emph{Proceedings of the 2020 {{Conference}} on {{Empirical
  Methods}} in {{Natural Language Processing}}: {{System Demonstrations}}},
  pages 38--45, 2020.

\bibitem[Baayen and Lieber(1991)]{baayen1991FIX}
Harald Baayen and Rochelle Lieber.
\newblock Productivity and {E}nglish {D}erivation: A {C}orpus-{B}ased {S}tudy.
\newblock \emph{Linguistics}, 29\penalty0 (5):\penalty0 801--844, 1991.

\bibitem[Gwet(2008)]{gwet2008}
Kilem~Li Gwet.
\newblock {Computing Inter-Rater Reliability and Its Variance in the Presence
  of High Agreement}.
\newblock \emph{British Journal of Mathematical and Statistical Psychology},
  61\penalty0 (1):\penalty0 29--48, 2008.

\bibitem[Liu et~al.(2023)Liu, Yuan, Fu, Jiang, Hayashi, and Neubig]{liu2023c}
Pengfei Liu, Weizhe Yuan, Jinlan Fu, Zhengbao Jiang, Hiroaki Hayashi, and
  Graham Neubig.
\newblock Pre-{T}rain, {{Prompt}}, and {{Predict}}: {{A Systematic Survey}} of
  {{Prompting Methods}} in {{Natural Language Processing}}.
\newblock \emph{ACM Computing Surveys}, 55\penalty0 (9):\penalty0 1--35,
  September 2023.

\bibitem[Crystal(1997)]{Crystal.1997}
David Crystal.
\newblock \emph{The {{Cambridge}} {E}ncyclopedia of the {{English}}
  {L}anguage}.
\newblock Cambridge University Press, Cambridge, UK, 1997.

\bibitem[Plag(2003)]{Plag.2003}
Ingo Plag.
\newblock \emph{Word-Formation in English}.
\newblock Cambridge University Press, Cambridge, UK, 2003.

\end{thebibliography}
}

\clearpage
\section*{Supporting Information} \label{si}

\subsection*{Derivative Statistics}

\begin{wraptable}{r}{0.6\linewidth}
\caption{Statistics of \ity{} and \ness{} derivatives for the 10 examined adjective classes in the Pile \citep{gao2020}, the corpus used to train \gpt{} \citep{wang2021a}. The total number of bases is 48,995. The values for token frequency are averaged across all word types belonging to a specific adjective class.}  
\label{tab:data-statistics}
\footnotesize
\centering
\begin{tabularx}{0.6\textwidth}{lr@{\hspace{0.75\tabcolsep}}rr@{\hspace{0.75\tabcolsep}}rr@{\hspace{0.75\tabcolsep}}rl}
\toprule
& \multicolumn{2}{c}{Type frequency} & \multicolumn{2}{c}{Token frequency} & \multicolumn{2}{c}{Hapaxes} \\
\cmidrule(lr){2-3}\cmidrule(lr){4-5}\cmidrule(lr){6-7}
Suffix & \ity{} & \ness{} & \ity{} & \ness{} & \ity{} & \ness{} \\ \midrule
\textit{-able}&11,081&1,034&3937.7&817.3&1,673&226\\
\textit{-al}&9,133&1,011&5904.9&172.1&2,078&251\\
\textit{-ar}&2,433&214&5833.7&10.3&451&59\\
\textit{-ed}&62&4,786&2.4&539.6&28&1,134\\
\textit{-ic}&6,215&617&4162.7&45.7&790&175\\
\textit{-ing}&2&1,600&1.0&1104.5&2&448\\
\textit{-ish}&0&1,502&0.0&397.0&0&437\\
\textit{-ive}&4,508&2,438&15075.8&3252.1&626&554\\
\textit{-less}&3&2,020&1.7&1159.8&1&506\\
\textit{-ous}&1,372&2,450&5453.1&2420.3&325&675\\
\bottomrule
\end{tabularx}
\end{wraptable}

We analyze the statistics of \ity{} and \ness{} derivatives in the Pile (Table~\ref{tab:data-statistics}).
We first focus on \emph{type frequency}, i.e., the number of different derivatives contained in the Pile. For most classes, there is a clear preference 
for either \ity{} or \ness{}, the only two exceptions being 
adjectives ending in \textit{-ive} and \textit{-ous}. For adjectives ending in the Germanic suffixes \textit{-ed}, \textit{-ing}, \textit{-ish}, and \textit{-less},
there is a particularly strong preference for \ness{}, although a few derivatives in \ity{} can be found
in the data. These statistics are similar to the results of a recent analysis based on dictionary data \citep{arndt-lappe2014}, indicating 
that the Pile provides a realistic picture of the variation between \ity{} and \ness{} in present-day English.

Next, we turn to \emph{token frequency}, i.e., the number of times individual derivatives occur in the Pile. We notice that 
the trends for type frequency are largely reflected by token frequency: in the case of adjective classes for which \ity{} derivatives have a higher 
type frequency than \ness{} derivatives, \ity{} derivatives also tend to have a higher average token frequency than \ness{} derivatives (and vice versa). The only exception is \textit{-ous}, where \ity{} has a lower 
type frequency but a higher average token frequency than \ness{}. This is 
due to a particularly large number of \ity{} derivatives in the high token frequency range: excluding
the top 5\% of derivatives with the highest token frequency, the average token frequency is higher for \ness{} (73.0) than \ity{} (15.1), 
in line with the type frequency trend for \textit{-ous}.

Finally, we examine a measure that linguistic 
scholarship has suggested to be particularly relevant for productivity \citep{baayen1991FIX, baayen1996}, 
specifically the number of \emph{hapaxes} (i.e., derivatives occurring only once in the Pile).
Here, the trends for individual adjective classes are similar to type frequency and 
token frequency, with the potential exception of \textit{-ive}, where the preponderance of \ity{} compared to \ness{}
is slightly less pronounced.

\subsection*{Nonce Adjectives}
\label{ap:pseudowords}

Table~\ref{tab:pseudowords} lists all nonce adjectives used in the paper.

\subsection*{Adjective Annotation}
\label{ap:annotation}

We collected human judgments from volunteers using the SoSciSurvey platform. Native speakers of English were recruited using snowball sampling and asked if they were willing to participate in a survey about derivational morphology. They were unaware of the exact goals of the study but knew that their data would be used for research purposes. In total, 22 participants took part in the survey.  All participants held at least an undergraduate college degree. The collection of adjective annotations was approved by the institutional review board of the Allen Institute for AI.

\begin{figure}[t!]
        \centering       
\begin{subfigure}[b]{\textwidth} 
        \centering       
            \includegraphics[width=0.75\textwidth]{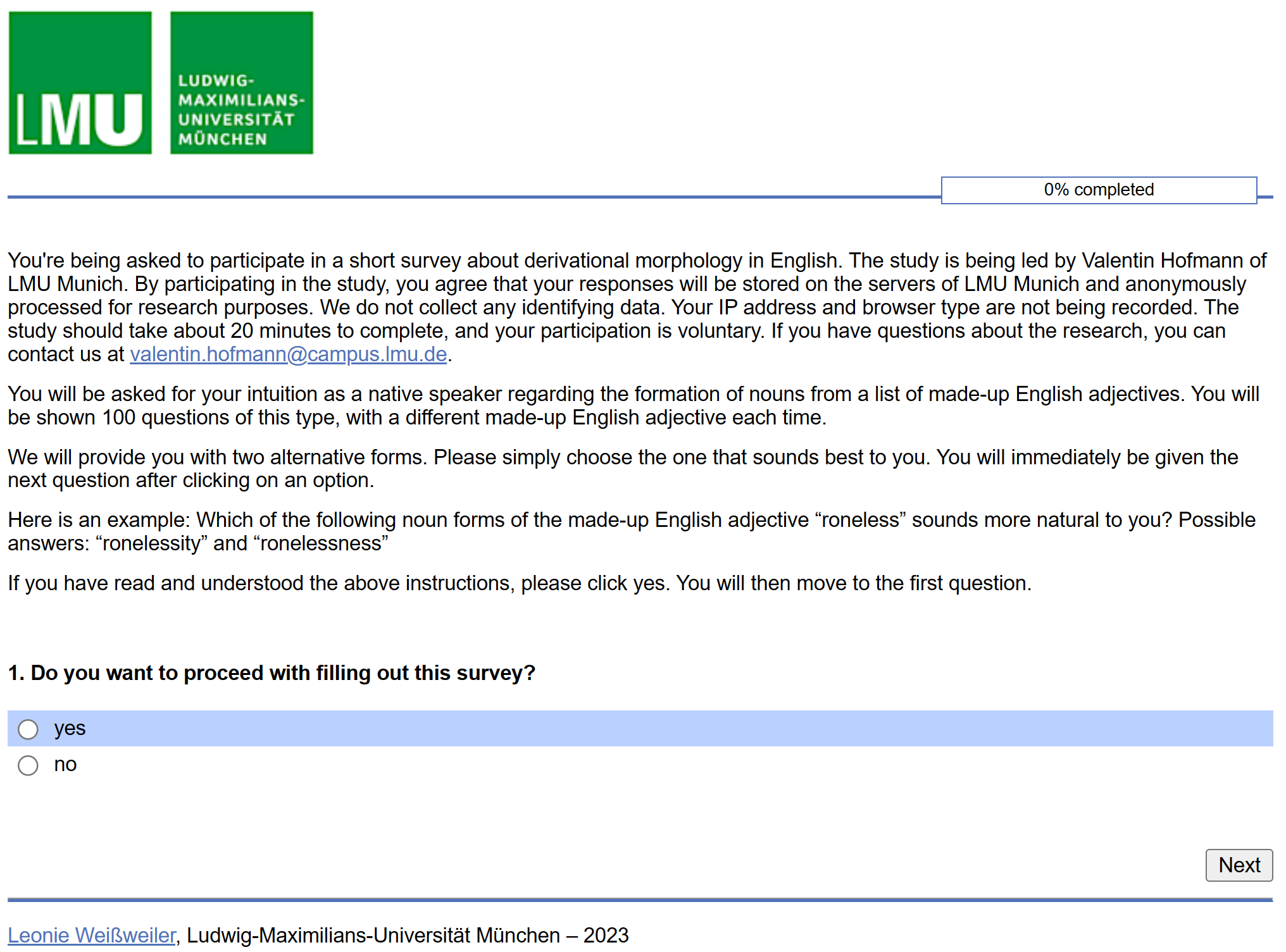}
            \caption[]%
            {{\small Introductory message}}    
            \label{fig:survey-intro}
        \end{subfigure}
 \begin{subfigure}[b]{\textwidth}  
         \centering       
         \includegraphics[width=0.75\textwidth]{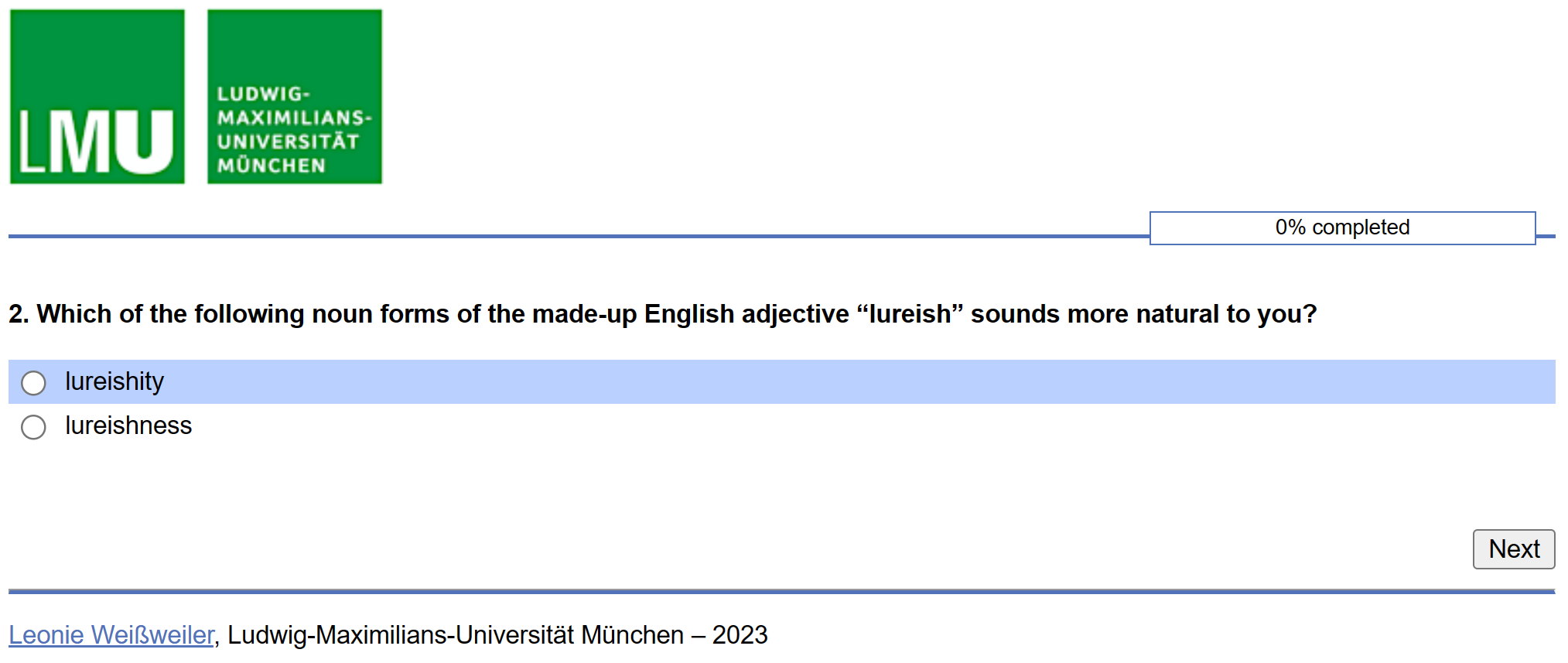} \caption[]{{\small Example question}}   \label{fig:survey-question}
 \end{subfigure}    
        \caption[]{Screenshots of the introductory message seen by participants of our survey (a) and an example question given to participants (b).}
\end{figure}

Each participant 
coded half of the nonce words (i.e., 100 nonce words), resulting 
in 11 annotations per nonce word. Participants were first shown an introductory message explaining the task as shown in Figure \ref{fig:survey-intro}. They were then given one of two survey versions, each with 100 nonce words that cycled through the four suffixes to avoid repetition. To reduce the total time necessary for completing the survey, participants were immediately shown the next question upon clicking one word. An example of a question is shown in Figure \ref{fig:survey-question}.

\subsection*{Annotator Variation}
\label{ap:annotator-variation}

\begin{wrapfigure}{r}{0.5\linewidth}
        \centering       
\begin{subfigure}[b]{0.21\textwidth}   
            \includegraphics[width=\textwidth]{figs/gptj_winner_ratio_unseen.pdf}
             \vspace{-5mm}
            \caption[]%
            {{\small \gpt{}}}    
            \label{fig:gptj-winner-ratio-unseen-repl}
        \end{subfigure}
 \begin{subfigure}[b]{0.21\textwidth}  \includegraphics[width=\textwidth]{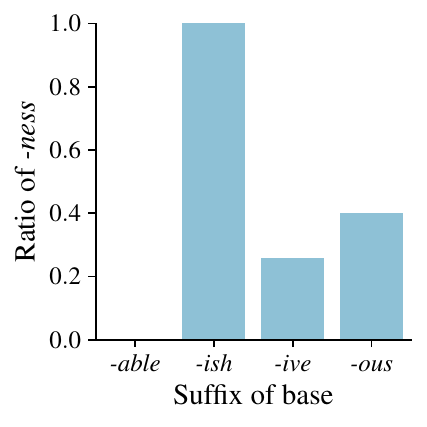} \vspace{-5mm} \caption[]{{\small Humans}}   \label{fig:humans-winner-ratio}
 \end{subfigure}    
        \caption[]{Distribution of preferred nominalization type (specifically, ratio of \ness{} derivatives) for unseen nonce adjectives,
        for GPT-J (a) and human annotators (b). The ratio is computed as the number of \ness{} predictions divided by the total number of predictions. Panel (a) replicates Figure \ref{fig:gptj-winner-ratio-unseen} from the main text for easier comparison.
        }
\end{wrapfigure}

\begin{wrapfigure}{r}{0.5\linewidth}
\centering       
\begin{subfigure}[b]{0.21\textwidth}  
\centering
            \includegraphics[width=\textwidth]{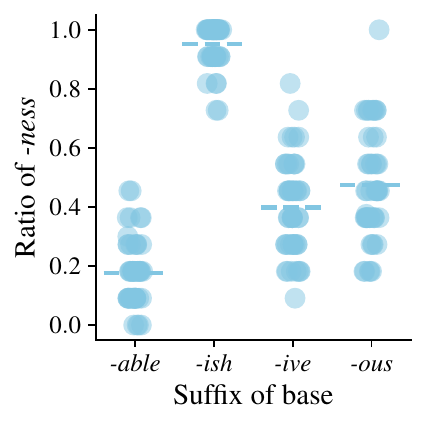}
             \vspace{-5mm}
            \caption[]%
            {{\small Base variation}}    
            \label{fig:humans-ity-ratio}
\end{subfigure}
\begin{subfigure}[b]{0.21\textwidth} \centering \includegraphics[width=\textwidth]{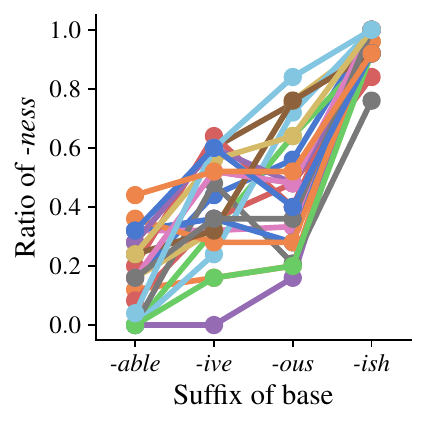} \vspace{-5mm} \caption[]{{\small Participant variation}}   \label{fig:participant-variation}
 \end{subfigure}    
        \caption[]{Variation in the derivative preferred by humans, shown separately for bases (a) and participants (b). In (a), each dot represents one base. In (b), each line represents the response pattern of one participant in our annotation study.}
\end{wrapfigure}

Figure~\ref{fig:humans-winner-ratio} plots for each tested adjective class
the ratio of bases for which participants overall preferred
\ness{} over \ity{}, i.e., more participants selected the \ness{} rather than the \ity{} derivative. 
There is a clear preference for 
\ity{} in the case of \textit{-able} and a clear preference for \ness{} in the case of \textit{-ish}.
For the two suffixes with a larger degree of competition, \textit{-ive} shows the expected pattern, 
with participants preferring \ity{} over \ness{} for the majority of bases, but \textit{-ous} 
shows a preference for \ity{}, which is different from its greater association with \ness{} in the Pile.
This can also be seen from the ratio of participants preferring \ness{} over \ity{} for individual 
bases (see Figure~\ref{fig:humans-ity-ratio}), which is on average smaller than 50\% for \textit{-able} (17.7\%), 
\textit{-ive} (39.8\%), and \textit{-ous} (47.5\%), and greater than 50\% only for \textit{-ish} (95.1\%). 
Figure~\ref{fig:humans-ity-ratio} also shows a high degree of variation between individual 
bases of a certain adjective class: e.g., for \textit{-ous}, there are bases for which participants 
clearly preferred \ity{} (e.g., 81.8\% preferred \ity{} for \textit{indaminous}), but there is also 
a base for which participants exclusively selected \ness{} (100\% preferred \ness{} for \textit{rebelorous}).

The human annotations vary also by participant: 
participants differed strongly in terms of how often they selected \ity{} or \ness{}
for each adjective class (see Figure~\ref{fig:participant-variation}). For example, 13 participants preferred  
\ity{} for \textit{-ous} bases, but nine participants preferred \ness{}.  The high degree of variation
is also 
reflected by an only small inter-annotator
agreement (IAA) of 0.335, measured using Fleiss' $\kappa$. However, measuring 
IAA on all bases hides the fact that IAA is substantially higher 
for \textit{-ish} (0.899) and \textit{-able} (0.587) 
than for \textit{-ive} (0.096) and \textit{-ous} (0.054), measured 
using Gwet's AC1 \citep{gwet2008}. We also notice
that there is a correlation between the responses given by individual participants for bases 
of different adjective classes, especially between \textit{-able} and \textit{-ive} (0.417),
and \textit{-ive} and \textit{-ous} (0.415), measured using Pearson's $r$. In other words, 
there is a moderate tendency for participants to prefer \ity{} or \ness{} for both 
\textit{-able} and \textit{-ive} bases (and similarly for \textit{-ive} and \textit{-ous} bases).

\subsection*{Prompts}
We want to test which of two derivatives --- the one ending in \textit{-ness} or the one ending in \textit{-ity} --- is preferred by a language model. To do so, we need to measure the probability that the language model assigns to the two competing forms. For example, we need to measure the probability that the language model assigns to \textit{sensitiveness}, and the probability that it assigns to \textit{sensitivity}. 
Language models such as \gpt{} and GPT-4 always assign probabilities to tokens \emph{given a sequence of preceding tokens}. Therefore, in order to measure the probability that a language model assigns to a specific derivative, we need to decide on what tokens to use as the preceding context. This is commonly referred to as \textit{prompting}, and the sequence of preceding tokens that is fed into the language model as \textit{prompt} \citep{liu2023c}. Properties of the prompt (e.g., the exact wording of a request) can substantially affect the language model predictions \citep{rae2022FIX}, which is why it has become common practice to examine several different prompts when analyzing the behavior of language models. Here, we use the following 12 prompts to measure the probabilities that \gpt{} assigns to the derivatives:
\begin{itemize}
\itemsep0em 
\item\textit{Nominalized adjective:}
\item\textit{Noun:}
\item\textit{The following is a nominalized adjective:}
\item\textit{The following is a noun:}
\item $b \rightarrow$
\item $b$ \textit{:}
\item $b$ \textit{-}
\item $b$
\item \textit{Adjective:} $b$ \textit{Nominalization:}
\item \textit{Form the nominalization of the given adjective.} $b \rightarrow$
\item \textit{Nominalize the given adjective.} $b \rightarrow$
\item \textit{Turn the given adjective into a noun.} $b \rightarrow$
\end{itemize}
As in the main text, $b$ here is a variable that refers to a base. For example, with the prompt \textit{Nominalized adjective:} and the base \textit{sensitive}, we measure the probability assigned to \textit{sensitivity} in the context \textit{Nominalized adjective: sensitivity} as well as the probability assigned to \textit{sensitiveness} in the context \textit{Nominalized adjective: sensitiveness}. The presented results are averaged across prompts; for example, to get GPT-J's match with the cognitive models in Table \ref{tab:comparison-cognitive-modes}, we calculate the match based on each of the 12 prompts and report the mean of these 12 scores.

We use the following prompts to measure the probabilities that \gpt{} assigns to the words in the vocabulary test:
\begin{itemize}
\itemsep0em 
\item \textit{Word:}
\item \textit{Real word:}
\item \textit{The following is a word:}
\item \textit{The following is a real word:}
\end{itemize}

\subsection*{GPT-4 Results}

\begin{wraptable}{r}{0.4\linewidth}
\caption{Match of rule-based  and exemplar-based models with GPT-4 on nonce adjectives.}  
\label{tab:gpt4}
\footnotesize
\centering
\begin{tabularx}{0.4\textwidth}{lr@{\hspace{0.95\tabcolsep}}rr@{\hspace{0.95\tabcolsep}}r}
\toprule
 & \multicolumn{2}{c}{MGL} & \multicolumn{2}{c}{GCM}\\ 
\cmidrule(lr){2-3}\cmidrule(lr){4-5}
Suffix  & Type & Token & Type & Token \\ \midrule
\textit{-able} & .960 & .960 & .960 & .960 \\
\textit{-ish} & 1.000 & 1.000 & 1.000 & 1.000 \\
\textit{-ive} & .400 & .480 & .440 & .500 \\
\textit{-ous} & .680 & .760  & .640 & .800 \\
\bottomrule
\end{tabularx}
\end{wraptable}

Table~\ref{tab:gpt4} shows the results of comparing the predictions of GPT-4 with the rule-based and examplar-based cognitive models examined in the main text (i.e., MGL and GCM, both in type-based and token-based variants).

\subsection*{Morphological Parse}
\label{ap:parse}

The parsability of words in the Hoosier lexicon is determined as follows. In a first step, we check whether a word is contained in CELEX \citep{baayen1995}, a lexical database that contains information about the morphological status of more than 50,000 English words. 16,417 words from the Hoosier lexicon are listed in CELEX. For the remaining 2,903 words, we determine the morphological status by means of a simple method from prior work \citep{hofmann2020a, hofmann-etal-2020-graph}: we test whether the beginning or end of words matches common prefixes/suffixes of the English language, and whether the remaining part of the word is a stem. To do so, we draw upon a list of 46 English prefixes and 44 English suffixes \citep{Crystal.1997}. As potential stems, we use all English words contained in CELEX. The algorithm is sensitive to morpho-orthographic rules of English \citep{Plag.2003}.

As a result of this procedure, 6,499 words from the Hoosier lexicon are classified as morphologically complex. The words are diverse in terms of the involved affixes: except for \textit{pseudo} and \textit{mini}, all affixes from the list mentioned above show up.

\begin{table*}[t]
\caption{Complete list of all used nonce adjectives.}
\label{tab:pseudowords}
    \centering
    \footnotesize
    {\itshape
    \begin{tabular}{llll}
    \toprule
-able & -ish & -ive & -ous \\ 
\midrule
actignable & badyish & atecusive & adodagious \\
anilicable & beavish & cogective & adupendous \\
anvastable & breyish & conovative & anoninous \\
chalinable & carmish & cormasive & aurtiguous \\
comfolvable & clangish & cuminitive & cazardous \\
compechable & clurlish & decertive & coivonous \\
condumable & cunkish & deflosive & creninous \\
contaitable & devevish & defrertive & dardulous \\
corgervable & direish & dejovative & dexarious \\
covornable & doutish & depulsive & erenymous \\
cresucable & dwaplish & dermasive & eretulous \\
enocutable & fadyish & dignitive & euphitious \\
expeaceable & fawkish & dimusitive & eutrigeous \\
expelocable & fevetish & exhauctive & faluminous \\
expernable & fevewish & expecative & fapturous \\
fispoceable & fevilish & extuctive & glamalous \\
fupeactable & frietish & gederative & glumonous \\
fusuperable & friquish & imimative & gluninous \\
imalatable & ghumpish & impuctive & gropenious \\
impalvable & gireish & indetative & hibeguous \\
inbeadable & goguish & nogensive & honoderous \\
inedifiable & higetish & nombasive & indaminous \\
infoustable & knarish & nonvuptive & iniragious \\
intoundable & laretish & nutensive & insicious \\
intountable & lureish & obsensive & lasavenous \\
inveicable & lurmish & pedititive & leamogous \\
irediocable & moguish & pedulsive & ligegious \\
mecoushable & peftish & pepulative & liratonous \\
parendable & preanish & pransitive & luticorous \\
peplaicable & prienish & prediasive & malicinous \\
praleckable & purerish & prititive & meglarious \\
preneckable & radyish & protrative & momogorous \\
prequakable & reckish & pumbative & mystuorous \\
previnable & redyish & recentive & nomeneous \\
previtable & rourfish & recumotive & oblicious \\
puneadable & shigeish & rejeptive & pecacious \\
pustameable & skierish & ruchontive & plalorous \\
redeptable & slarish & seceptive & poncorous \\
rempadable & slownish & sejensive & prolacious \\
retaleable & slundish & serposive & ralygerous \\
sempoivable & slungish & submiative & ravarious \\
swimitable & snoulish & submictive & reamorous \\
tegornable & sonkish & submistive & rebelorous \\
unaclerable & tivilish & sumpertive & slaicitous \\
unalintable & turgeish & sumurative & suspibious \\
undeperable & wabyish & suprective & tefigious \\
unutintable & waguish & tecensive & trospurous \\
unvatrable & wainish & tendusive & undicitous \\
unvediable & wawkish & tredictive & vexuteous \\
utililable & woungish & vederative & vombageous \\
\bottomrule
\end{tabular}}
\end{table*}

\end{document}